\documentclass[10pt,twocolumn,letterpaper]{article} % For LaTeX2e
\usepackage{cvpr}
\usepackage{times}
\def\confName{CVPR}\def\confYear{2027}

\usepackage{amsmath,amsfonts,bm}

\def\eqref#1{equation~\ref{#1}}
\def\plaineqref#1{\ref{#1}}
\def\1{\bm{1}}

\DeclareMathAlphabet{\mathsfit}{\encodingdefault}{\sfdefault}{m}{sl}
\SetMathAlphabet{\mathsfit}{bold}{\encodingdefault}{\sfdefault}{bx}{n}

\newcommand{\E}{\mathbb{E}}

\newcommand{\KL}{D_{\mathrm{KL}}}

\usepackage[hidelinks]{hyperref}

\usepackage{amsmath,amssymb,mathtools,bm}
\usepackage{booktabs}
\usepackage{graphicx}
\usepackage{multirow}
\usepackage{placeins}
\usepackage{xspace}
\usepackage{tikz}
\usepackage{xcolor}

\usetikzlibrary{arrows.meta,positioning,fit,calc}

\newcommand{\paper}{BridgeMatch\xspace}

\newcommand{\N}{\mathcal{N}}
\newcommand{\I}{\mathbf{I}}
\newcommand{\X}{\mathbf{X}}
\newcommand{\Y}{\mathbf{Y}}
\newcommand{\C}{\mathbf{C}}
\newcommand{\F}{\mathbf{F}}
\newcommand{\U}{\mathcal{U}}

\newcommand{\clip}{\operatorname{clip}}

\definecolor{diffblue}{RGB}{79,129,189}
\definecolor{liftgreen}{RGB}{112,173,71}
\definecolor{floworange}{RGB}{237,125,49}
\definecolor{sbpurple}{RGB}{112,48,160}
\definecolor{lightgray}{RGB}{245,245,245}

\definecolor{cvprblue}{rgb}{0.21,0.49,0.74}
\definecolor{mycolor}{HTML}{40405C}
\definecolor{titlegray}{HTML}{585858}
\definecolor{mycolor}{HTML}{008F7A}
\title{BridgeMatch: Conditional Transport Bridges in Matching Matrix Space for 3D Deformable Registration}

\author{
Qianliang Wu$^{1}$ \quad Yaqing Ding$^{4}$ \quad Guangwei Gao$^{2}$\\
Weiping Ding$^{1}$ \quad Jin Xie$^{3}$ \quad Jian Yang$^{2}$\\[3pt]
$^{1}$Nantong University \quad $^{2}$Nanjing University of Science and Technology\\
$^{3}$Nanjing University \quad $^{4}$Southeast University
}

\begin{document}
\hfuzz=0pt\vfuzz=0pt

\maketitle

\begin{abstract}
Non-rigid point cloud registration depends on accurate correspondences under deformation and partial visibility. Existing coarse-to-fine methods use coarse matches to select regions for fine matching, but incorrect selection can remove valid correspondences before refinement. We introduce \paper, a conditional transport framework that connects coarse correspondence estimation to high-resolution matching. A diffusion model first estimates a coarse matching matrix. We lift the complete estimate to a fine-level source matrix and refine its scores with a geometry-conditioned bridge. We consider two bridge formulations: a deterministic endpoint-parameterized flow and a stochastic paired Brownian bridge with endpoint and noise prediction. At each step, the current matrix provides an alignment for the next prediction. Both models reuse backbone features throughout inference. Experiments on 4DMatch and 4DLoMatch show improved correspondence and non-rigid registration accuracy. The models also transfer to CAPE and DeepDeform without target-domain training. Ablations examine the roles of candidate coverage, source initialization, and refinement budget.
\end{abstract}
\section{Introduction}
\label{sec:intro}
Non-rigid registration aligns point clouds whose relative motion cannot be described by a single rigid transformation. It is used in dynamic reconstruction~\cite{prokudin2023dynamic}, medical imaging~\cite{zou2022review,yang2025boundary}, and deformable-object manipulation~\cite{shi2022robocraft,lin2023planning}. Reliable correspondences provide constraints for estimating the deformation. Finding them is difficult when scans have limited overlap, similar local structures, or large changes in shape.

Learned matching methods extract point features at multiple resolutions and use geometric context to estimate correspondences~\cite{li2022lepard,qin2022geometric,yu2023rotation}. Coarse features capture a wider spatial context, while fine features retain local detail. Coarse-to-fine pipelines often connect these resolutions by selecting high-confidence coarse pairs and matching within their associated regions~\cite{yu2021cofinet,qin2022geometric}. This selection reduces the search space, but a correct fine-level pair can be lost if its coarse pair receives a low score. Subsequent refinement is then restricted to the surviving candidates.

Coarse and fine matching provide different information. Coarse estimates capture spatial context but merge points with distinct local geometry. Fine features can resolve these differences, provided that the relevant candidates remain available. We therefore use the complete coarse estimate as the starting point for fine matching. The fine stage can then revise a low coarse score while keeping the corresponding pair available.

\begin{figure*}[!t]
	\centering
		\includegraphics[width=\textwidth]{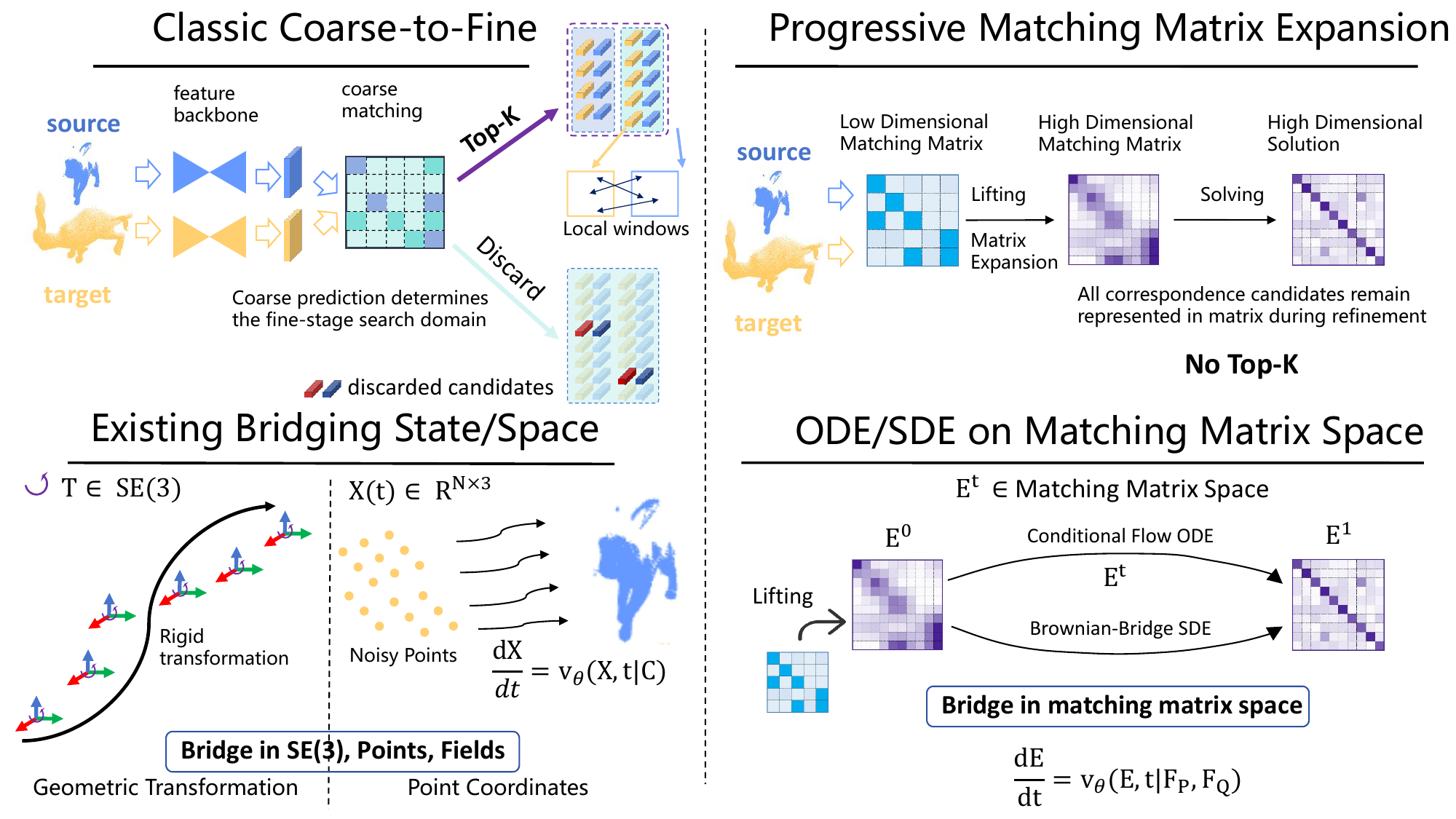}
	\caption{Comparison with existing paradigms. Traditional coarse-to-fine pipelines restrict fine matching through Top-$K$ pruning, while our progressive expansion retains all candidates by lifting the complete coarse matrix. Unlike bridges in coordinate or transformation space, \paper transports the matching matrix ($\mathbf{E}_t$) through a conditional-flow ODE or a Brownian-bridge SDE.}
	\label{fig:difference_method}
	
\end{figure*}

\begin{figure*}[!t]
	\centering
	\includegraphics[width=0.9\textwidth]{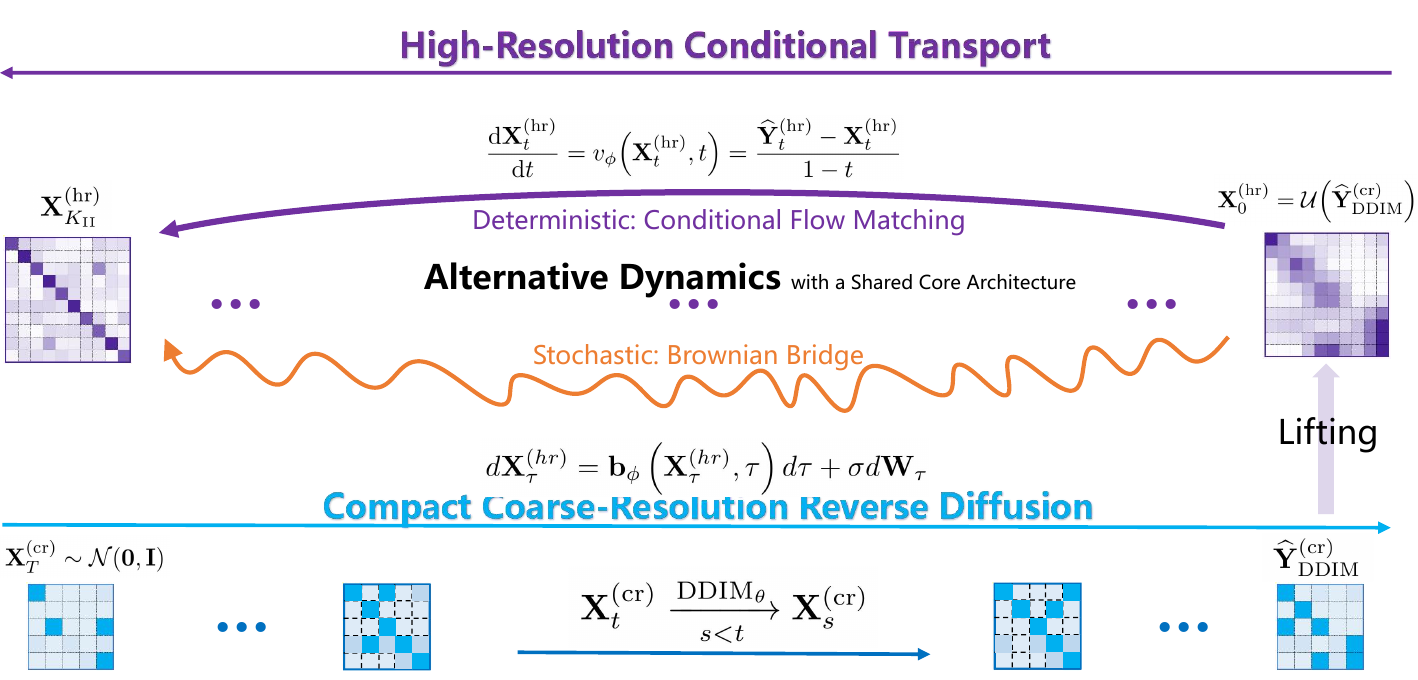}
	\caption{\paper architecture. Stage~I estimates coarse matches by diffusion; lifting supplies the initial matrix for Stage~II. Both Stage-II implementations use the same endpoint-predictor architecture; the stochastic model adds a noise head to construct the forward drift.}
	\label{fig:framework}
\end{figure*}

We propose \paper, a two-stage model that learns a conditional path from a lifted coarse estimate to the high-resolution target matrix. Stage~I uses diffusion to predict coarse correspondences. A lifting operation copies each coarse score to the fine-level pairs associated with its parent points. Stage~II starts from this structured matrix and refines its entries with fine-level geometry. No candidate is removed during lifting. Figure~\ref{fig:difference_method} illustrates this change in representation, and Fig.~\ref{fig:framework} shows the architecture.

The fine stage uses the current matching matrix to estimate an alignment, transform the source points, and condition the next endpoint prediction. We implement this process with two conditional bridges. The deterministic model predicts the endpoint along a straight path and uses it to define a flow-matching ODE~\cite{lipman2023flowmatching,liu2022flow}. The stochastic model trains on a paired Brownian path and combines endpoint and noise predictions to construct the SDE drift~\cite{tong2024simulation}. Each variant is trained separately with the coarse stage. At inference, backbone features are computed once; the geometric conditioning and matching predictions are updated at each step.

We evaluate correspondence estimation and downstream registration on 4DMatch and 4DLoMatch. The stochastic model obtains $82.41\%$ NFMR and $78.79\%$ IR on 4DLoMatch. Both variants improve registration over Diff-Reg with the same GraphSCNet solver and transfer to CAPE and DeepDeform without adaptation. Candidate-masking experiments measure how many discarded correct pairs can be recovered by full-matrix refinement. Further experiments compare source initialization, update rules, and computational cost.

The contributions are:
\begin{itemize}
 \item A two-stage matching model that connects coarse diffusion to high-resolution conditional transport through hierarchy-preserving lifting.
 \item Deterministic endpoint-CFM and stochastic paired Brownian-bridge formulations for refining the complete fine-level matching matrix.
 \item Experiments on non-rigid matching, registration, and cross-dataset transfer, with controlled studies of candidate restriction and inference dynamics.
\end{itemize}

\section{Related Work}
\label{sec:related}
\paragraph{Point Cloud Registration.}
Recent registration methods address feature learning, correspondence consistency, and deformation estimation. PARE-Net uses position-aware rotation-equivariant features to generate pose hypotheses from individual correspondences~\citep{yao2024parenet}. CAST focuses cross-attention on candidate regions and uses geometric consistency in self-attention~\citep{huang2024cast}. 

For non-rigid alignment, NDP decomposes motion through an MLP pyramid with increasing positional frequencies~\citep{li2022non}; GraphSCNet rejects outliers using local spatial consistency on a deformation graph~\citep{qin2023deep}. SyNoRiM synchronizes learned functional maps to enforce cycle consistency across scans~\citep{huang2022multiway}. OAR uses adaptive correntropy to reduce occlusion errors and local reconstruction to constrain unmatched regions~\citep{zhao2025occlusion}. ERNet predicts deformation graph nodes and refines their trajectories in temporal windows~\citep{He_2025_ICCV}. \citet{Chen_2026_CVPR} learn topology-aware propagation and codebook-enhanced superpoint features for deformation robustness. AniSym-Net couples anisotropic shape--motion fields with symplectic constraints to handle deformation and occlusion~\citep{wang2025anisym}. RGGT combines pretrained generative features with correspondence--reconstruction supervision for rigid and non-rigid registration~\citep{zhengrggt}. UniCorrn shares transformer weights across 2D/3D matching tasks, with separate appearance and position streams~\citep{Goswami_2026_CVPR}.

\paragraph{Generative Registration.}
Generative methods perform registration by refining matching matrices, poses, or point coordinates. Diff-PCR searches doubly stochastic matrix space through iterative denoising~\citep{wu2023diff}. Diff-Reg supports 3D and 2D--3D matching with a lightweight denoiser that reuses backbone features across steps~\citep{wu2024diff}. ODIN alternates overlap discovery with correlation-matrix denoising~\citep{jin2024multiway}. 

FUSER predicts multiview poses jointly; FUSER-DF refines them by diffusion in $\mathrm{SE}(3)^N$~\citep{jiang2026fuser}. Rectified Point Flow learns point-wise velocities for registration and shape assembly, then recovers poses from transported points~\citep{sun2025rectified}. Register Any Point scales this coordinate-transport approach to multiview scenes with test-time rigidity enforcement~\citep{pan2026register}. ODin generates aligned human point clouds with fixed point ordering to preserve template semantics~\citep{masiero2026ordered}.

\paragraph{Flow Matching and Stochastic Bridges.}
Flow matching regresses conditional vector fields without simulating training trajectories~\citep{lipman2023flowmatching}; rectified flow fits linear interpolations and uses reflow to straighten sampling paths~\citep{liu2022flow}. Riemannian flow matching constructs target fields from geometry-aware premetrics~\citep{chen2024riemannian}. GaussianAnything separates geometry and appearance generation through cascaded flows in point-structured latent space~\citep{lan2025gaussiananything}. SGMatch combines semantic local attention with flow-based feature transport regularization for non-rigid correspondence~\citep{ye2026sgmatch}. 

For stochastic transport, \citet{tong2024simulation} combine score and flow matching with entropy-regularized endpoint coupling to learn Schr{\"o}dinger bridges without trajectory simulation. P2P-Bridge learns bridges from noisy to clean point coordinates~\citep{vogel2024p2pbridge}; BridgeShape connects incomplete and complete shapes in a depth-enhanced VQ-VAE latent space~\citep{kong2026bridgeshape}.

\section{Method}
\label{sec:method}

\subsection{Method Overview}
\label{sec:overview}

Given source and target point clouds $\mathcal{P}=\{\mathbf{p}_i\}_{i=1}^{N}$ and $\mathcal{Q}=\{\mathbf{q}_j\}_{j=1}^{M}$, we estimate a soft matching matrix for downstream non-rigid registration. A hierarchical backbone produces features $\F_s^{(r)},\F_t^{(r)}$ at coarse ($r=\mathrm{cr}$) and high ($r=\mathrm{hr}$) resolutions. Binary matching matrices $\Y^{(r)}\in\{0,1\}^{N_r\times M_r}$ provide supervision, with positive entries marking ground-truth correspondences.

Stage~I denoises a compact matrix from Gaussian noise to estimate global correspondences. We lift this estimate to form the source matrix $\X_0$. Stage~II then refines all candidate scores using fine-level geometry (Fig.~\ref{fig:framework}).

\subsection{Stage I: Coarse Correspondence Diffusion}
\label{sec:l1}

Forward diffusion adds Gaussian noise to the coarse target matrix. With cumulative signal coefficient $\bar\alpha_k$, the noisy matrix at step $k$ is
\begin{equation}
	\X_k^{(\mathrm{cr})}=\sqrt{\bar\alpha_k}\,\Y^{(\mathrm{cr})}
	+\sqrt{1-\bar\alpha_k}\,\bm\epsilon,
	\quad \bm\epsilon\sim\N(\mathbf{0},\I).
	\label{eq:l1_forward}
\end{equation}
A time-conditioned matching transformer uses this noisy matrix and coarse features to predict the clean endpoint:
\begin{equation}
	\widehat{\Y}_{0,k}^{(\mathrm{cr})}
	=D_{\theta}\!\left(\X_k^{(\mathrm{cr})},k,
	\F_s^{(\mathrm{cr})},\F_t^{(\mathrm{cr})}\right).
	\label{eq:l1_endpoint}
\end{equation}
A bounded confidence view of $\X_k^{(\mathrm{cr})}$ provides the geometric conditioning used by the transformer.

During inference, deterministic DDIM~\cite{song2020denoising} starts from $\X_T^{(\mathrm{cr})}\sim\N(\mathbf{0},\I)$ and produces $\widehat{\Y}_{\mathrm{DDIM}}^{(\mathrm{cr})}$.

\subsection{Lifting to High Resolution}
\label{sec:lifting}

We detach the coarse prediction and transfer its entries using the point hierarchy. Let $\mathbf{P}_s\in\{0,1\}^{N_{\mathrm{hr}}\times N_{\mathrm{cr}}}$ and $\mathbf{P}_t\in\{0,1\}^{M_{\mathrm{hr}}\times M_{\mathrm{cr}}}$ be parent-assignment matrices, each with one nonzero entry per row. The lifted source is
\begin{equation}
	\X_0=\U(\widehat{\Y}^{(\mathrm{cr})})
	=\mathbf{P}_s\widehat{\Y}^{(\mathrm{cr})}\mathbf{P}_t^{\top}.
	\label{eq:lifting}
\end{equation}
We copy each coarse entry to all high-resolution pairs with the same parent points. This retains the complete candidate space. During refinement, fine-level features distinguish candidates within each block.

\subsection{Stage-II Conditional Paths and Endpoint Predictor}
\label{sec:shared_bridge}

Stage~II learns from intermediate states between the source matrix and high-resolution target. During training, a curriculum selects $\X_0$ by varying coarse predictions and gradually removing ground-truth mixing. At inference, $\X_0$ is the lifted DDIM estimate. Conditioned on point geometry and backbone features, the training path is
\begin{equation}
\begin{aligned}
\X_t^{(d)}&=(1-t)\X_0+t\Y^{(\mathrm{hr})}+\rho_d(t)\bm\epsilon,\\
\rho_d(t)&=\begin{cases}
0, & d=\mathrm{ODE},\\
\sigma_B\sqrt{t(1-t)}, & d=\mathrm{SDE}.
\end{cases}
\end{aligned}
\label{eq:shared_stage2_path}
\end{equation}
Here, $d\in\{\mathrm{ODE},\mathrm{SDE}\}$ selects the path, and $\bm\epsilon\sim\N(\mathbf{0},\I)$. Below we omit $\mathrm{hr}$: $\X_t$ denotes the current state and $\widehat{\Y}_t$ the predicted endpoint. In Fig.~\ref{fig:framework}, $\X_{K_{\mathrm{II}}}^{(\mathrm{hr})}$ is the state after $K_{\mathrm{II}}$ solver steps.

The endpoint predictor first forms a bounded view $\mathbf{C}_t$ of the state. Endpoint-CFM uses its bounded state directly; the Brownian branch uses $\mathbf{C}_t=\clip(\X_t,0,1)$. After masking invalid pairs, Sinkhorn normalization gives assignment weights $\mathbf{A}_t$. Soft Procrustes then estimates
\begin{equation}
	(\mathbf{R}_t,\bm\tau_t)=
	\arg\min_{\mathbf{R}\in\mathrm{SO}(3),\bm\tau}
	\sum_{ij} A_{t,ij}
	\|\mathbf{R}\mathbf{p}_i+\bm\tau-\mathbf{q}_j\|_2^2.
	\label{eq:soft_proc}
\end{equation}
We use this transform to condition the next prediction on the current alignment. The transformer processes the warped source $\widetilde{\mathbf{p}}_{i,t}=\mathbf{R}_t\mathbf{p}_i+\bm\tau_t$, target geometry, high-resolution features, and a sinusoidal time embedding:
\begin{equation}
	\widehat{\Y}_t
	=H_{\phi}\!\left(\mathbf{C}_t,t,\F_s^{(\mathrm{hr})},\F_t^{(\mathrm{hr})},
	\widetilde{\mathcal{P}}_t,\mathcal{Q}\right).
	\label{eq:shared_endpoint}
\end{equation}
The matrix $\mathbf{C}_t$ also biases the matching logits. Backbone features are computed once; the geometric alignment and endpoint prediction are updated at each step. In the stochastic branch, the noise head $G_{\psi}$ and SDE update receive the unbounded $\X_t$, while clipping is confined to endpoint conditioning.

\subsection{Stage II-A: Deterministic Endpoint-CFM Solver}
\label{sec:method_fm}

The deterministic model sets $\rho_{\mathrm{ODE}}(t)=0$ in Eq.~\plaineqref{eq:shared_stage2_path}. The endpoint prediction defines the velocity
\begin{equation}
	v_{\phi}(\X_t,t)
	=\frac{\widehat{\Y}_t-\X_t}{1-t}.
	\label{eq:endpoint_velocity}
\end{equation}
Only a small fraction of matching-matrix entries are positive. We supervise the endpoint with focal loss to emphasize misclassified entries and compute the velocity using Eq.~\plaineqref{eq:endpoint_velocity}. On the straight training path, squared velocity error equals squared endpoint error weighted by $(1-t)^{-2}$.

Inference begins at the lifted DDIM matrix and solves
\begin{equation}
	\frac{d\X_t}{dt}=v_{\phi}(\X_t,t),
	\qquad \X_{0}=\U(\widehat{\Y}_{\mathrm{DDIM}}^{(\mathrm{cr})}).
	\label{eq:fm_inference}
\end{equation}
We use explicit Euler steps with bounded updates and evaluate the field before the singular endpoint $t=1$. The final Euler update equals the clipped endpoint prediction.

\subsection{Stage II-B: Stochastic Paired Brownian-Bridge Solver}
\label{sec:method_sb}

For the Brownian path, let $s_t=\sqrt{t(1-t)}$. The noise head $G_{\psi}$ predicts $\widehat{\bm\epsilon}_t$ from the matching state, source matrix, time, and shared geometric features. Combined with the endpoint prediction, it gives the drift
\begin{equation}
	b_{\phi}(\X_t,t)
	=\widehat{\Y}_t-\X_0
	-\sigma_B\frac{t}{s_t}\widehat{\bm\epsilon}_t.
	\label{eq:sb_drift_explicit}
\end{equation}
The sampler follows
\begin{equation}
	\mathrm{d}\X_t
	=b_{\phi}(\X_t,t)\,\mathrm{d}t
	+\sigma_B\,\mathrm{d}\mathbf{W}_t,
	\label{eq:sb_forward_sde}
\end{equation}
where $\mathbf{W}_t$ is a standard matrix-valued Wiener process. At inference, we initialize $\X_{\varepsilon}=\X_0+\sigma_B\sqrt{\varepsilon(1-\varepsilon)}\,\mathbf z$, where $\mathbf z\sim\N(\mathbf 0,\I)$. This matches the training path's noise variance at $\varepsilon$ and centers the state at the available source $\X_0$. Euler--Maruyama steps over $[\varepsilon,1-\varepsilon]$ are followed by a final endpoint prediction. With exact endpoint and noise estimates on the paired training path, the drift reduces to $(\Y-\X_t)/(1-t)$.

Each run jointly trains Stage~I and one Stage-II branch with focal correspondence and geometric motion losses. The stochastic branch averages noise errors separately over positive and negative pairs, giving each class equal total weight. Endpoint and noise predictions define the drift through Eq.~\plaineqref{eq:sb_drift_explicit}. Coarse predictions are detached before lifting; both correspondence losses update the shared backbone.

\section{Experiments}
\label{sec:experiments}

\subsection{Protocol}

\paragraph{Datasets.}
We evaluate matching on 4DMatch and 4DLoMatch~\cite{li2022lepard}, which contain partial point-cloud pairs with non-rigid motion. A $45\%$ overlap boundary separates the test splits, with 4DLoMatch covering lower overlap. Downstream registration uses the filtered 4DMatch-F/4DLoMatch-F splits~\cite{li2022non,qin2023deep}.

\paragraph{Metrics.}
We report non-rigid feature matching recall (NFMR) and inlier ratio (IR) at a $4$\,cm threshold. NFMR measures ground-truth match recovery through interpolation from predicted matches. IR is the fraction of predicted matches correct under the ground-truth warp. For registration, end-point error (EPE) averages distances between predicted and ground-truth warped points. Strict and relaxed accuracy (AccS/AccR) measure the fractions of points meeting their absolute-or-relative error thresholds. Outlier ratio (OR) measures the fraction of points with relative error above $30\%$. EPE is in meters; other metrics are percentages. Lower EPE/OR and higher accuracy or recall indicate better performance.

\paragraph{Zero-shot protocol.}
Directed test pairs come from MPC-CAPE and MPC-DD, the preprocessed CAPE and DeepDeform datasets released with SyNoRiM~\cite{huang2022multiway}. Diff-Reg and our models use 4DMatch-only checkpoints and the same fixed 4DMatch-trained GraphSCNet on both datasets, without target-domain adaptation. The matched DeepDeform evaluation also includes Lepard. Stochastic results are averaged over sampling seeds $\{0,1,2\}$; deterministic matchers run once unless noted. The symbols $^*$ and $^\dagger$ mark target-domain training and different evaluation protocols, respectively. AccS/AccR thresholds are $(2\,\mathrm{cm},5\%)/(5\,\mathrm{cm},10\%)$ on CAPE and $(2.5\,\mathrm{cm},2.5\%)/(5\,\mathrm{cm},5\%)$ on DeepDeform. Metrics are averaged over test pairs.

\paragraph{Training and inference settings.}
We use Lepard's KPFCN hierarchy and train for 60 epochs with batch size 2 and SGD (learning rate $0.015$, momentum $0.93$, weight decay $10^{-6}$, exponential decay $0.95$, gradient clipping at 5). Both solvers default to 5 coarse DDIM and 20 fine bridge steps; SDE uses $\sigma_B=0.1$ and $\varepsilon=0.02$. We extract correspondences by mutual nearest matching with confidence threshold $0.2$. The ODE and SDE models are trained separately with matched epochs and step counts but different paths, auxiliary heads, and losses.

\subsection{Non-Rigid Correspondence and Registration}

\begin{figure*}[p]
	\centering
	\setlength{\tabcolsep}{1pt}
	
	\begin{tabular}{@{}cc cc cc cc cc@{}}
		\multicolumn{2}{c}{GeoTR} &
		\multicolumn{2}{c}{RoITr} &
		\multicolumn{2}{c}{Diff-Reg} &
		\multicolumn{2}{c}{Ours$^{\mathrm{ODE}}$} &
		\multicolumn{2}{c}{Ours$^{\mathrm{Stochastic}}$} \\
		
		\includegraphics[width=0.094\textwidth,height=1.40cm]
		{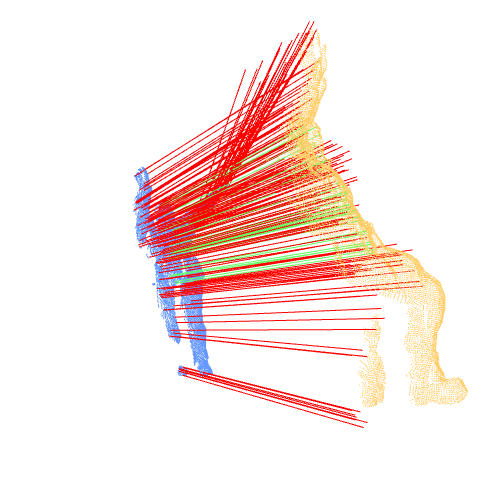} &
		\includegraphics[width=0.094\textwidth,height=1.40cm]
		{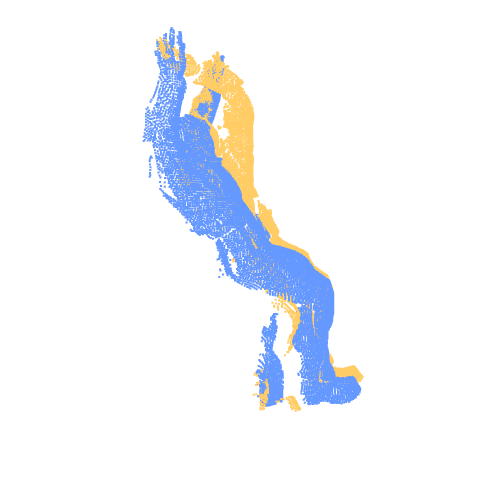} &
		\includegraphics[width=0.094\textwidth,height=1.40cm]
		{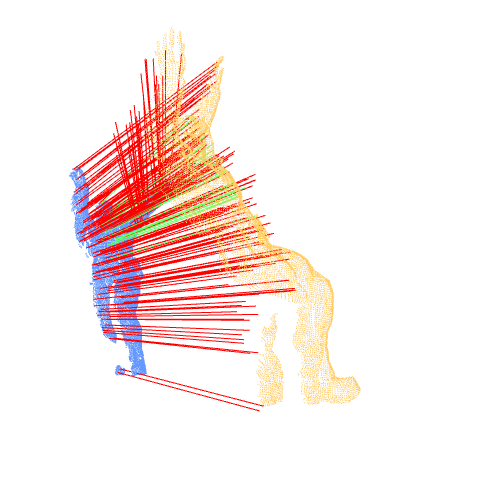} &
		\includegraphics[width=0.094\textwidth,height=1.40cm]
		{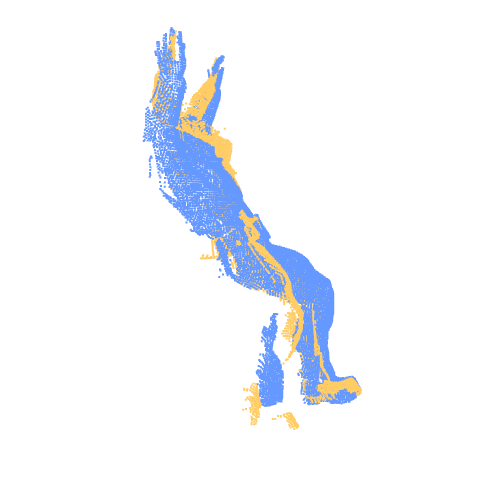} &
		\includegraphics[width=0.094\textwidth,height=1.40cm]
		{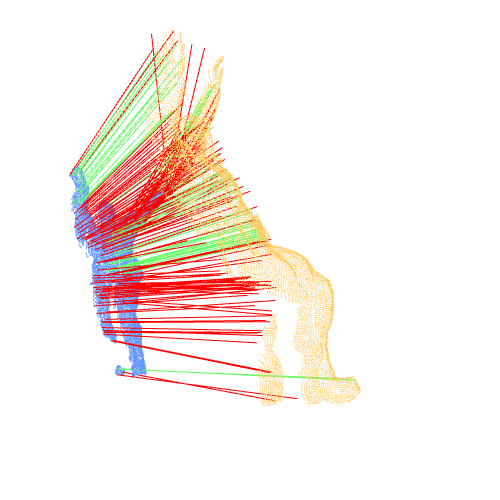} &
		\includegraphics[width=0.094\textwidth,height=1.40cm]
		{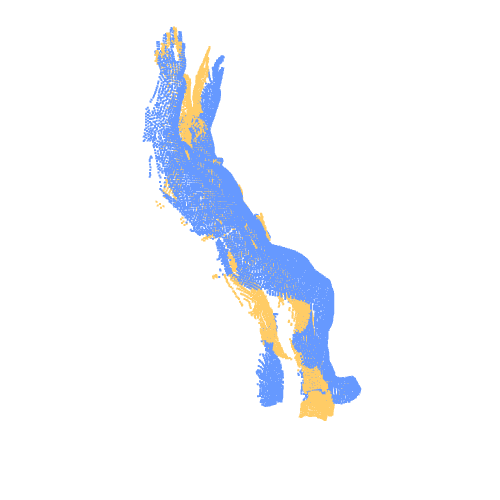} &
		\includegraphics[width=0.094\textwidth,height=1.40cm]
		{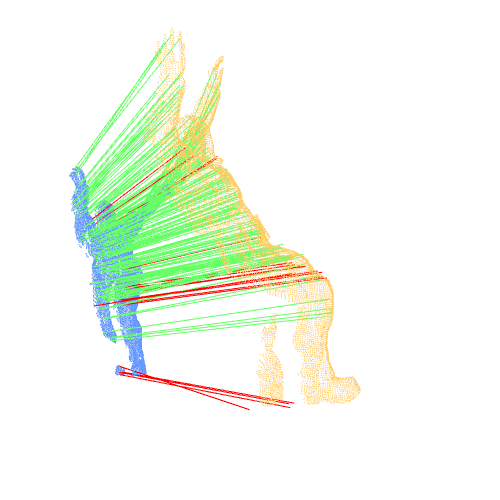} &
		\includegraphics[width=0.094\textwidth,height=1.40cm]
		{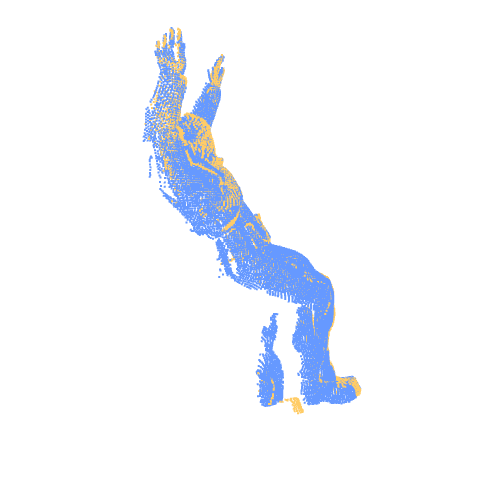} &
		\includegraphics[width=0.094\textwidth,height=1.40cm]
		{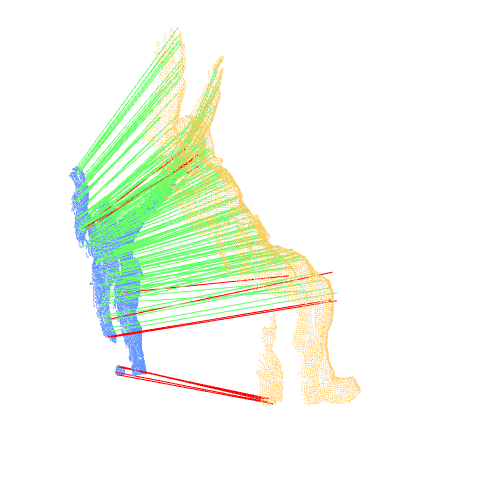} &
		\includegraphics[width=0.094\textwidth,height=1.40cm]
		{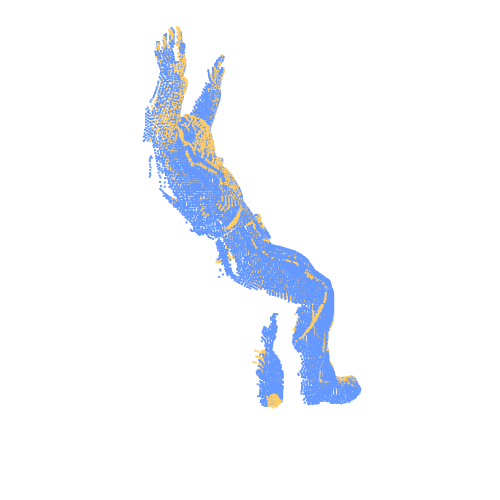} \\
		
		\includegraphics[width=0.094\textwidth,height=1.40cm]
		{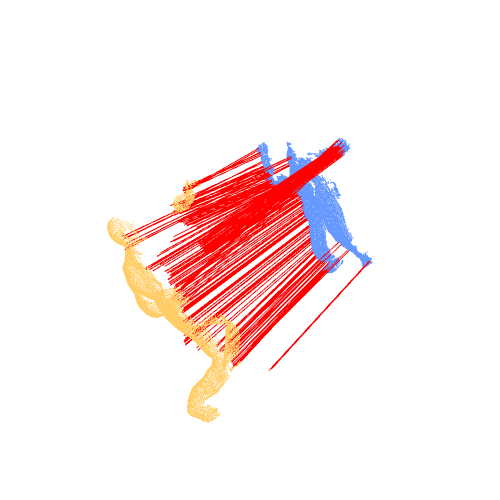} &
		\includegraphics[width=0.094\textwidth,height=1.40cm]
		{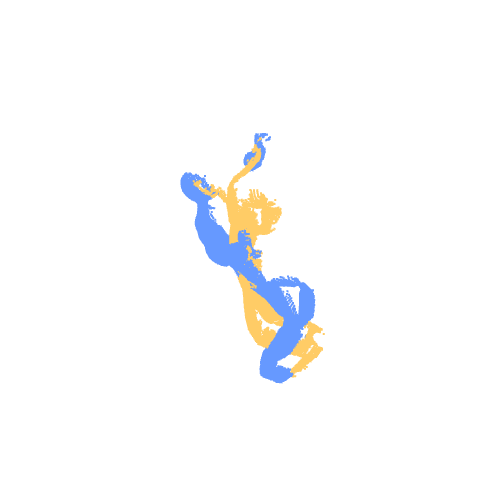} &
		\includegraphics[width=0.094\textwidth,height=1.40cm]
		{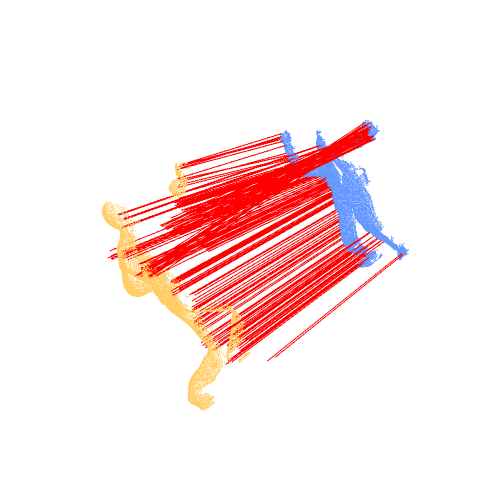} &
		\includegraphics[width=0.094\textwidth,height=1.40cm]
		{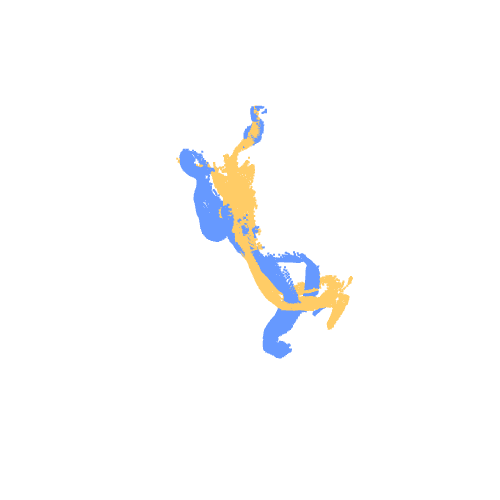} &
		\includegraphics[width=0.094\textwidth,height=1.40cm]
		{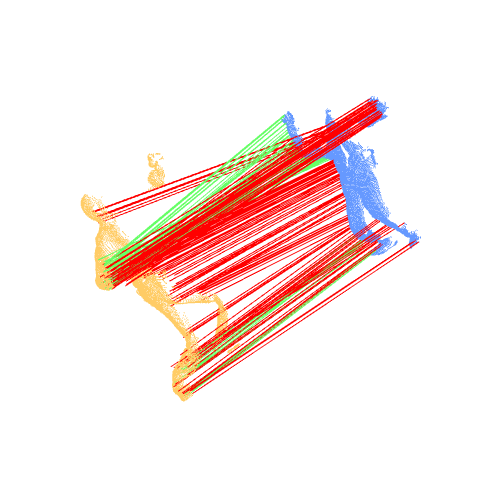} &
		\includegraphics[width=0.094\textwidth,height=1.40cm]
		{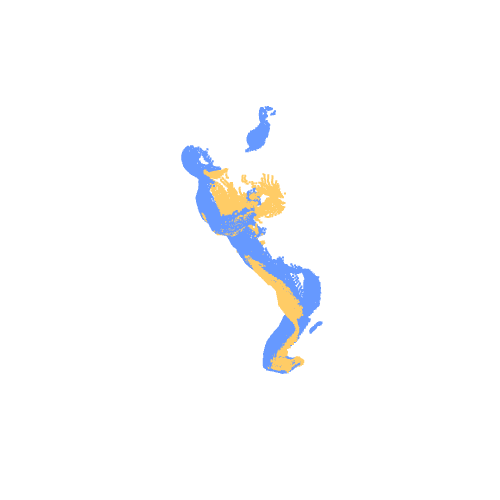} &
		\includegraphics[width=0.094\textwidth,height=1.40cm]
		{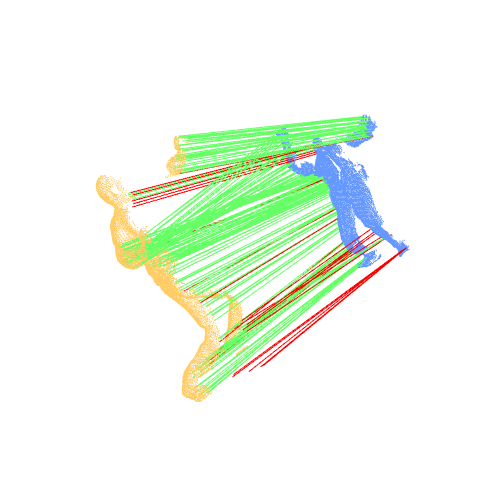} &
		\includegraphics[width=0.094\textwidth,height=1.40cm]
		{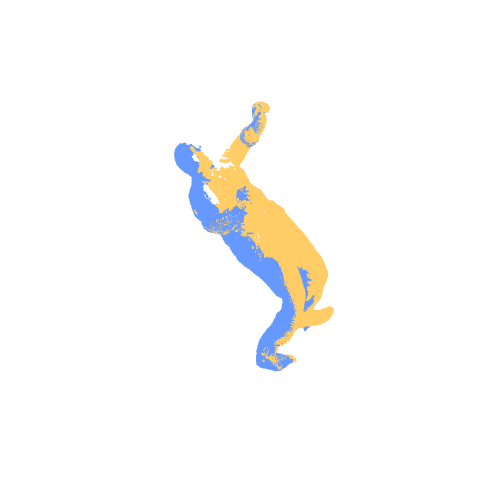} &
		\includegraphics[width=0.094\textwidth,height=1.40cm]
		{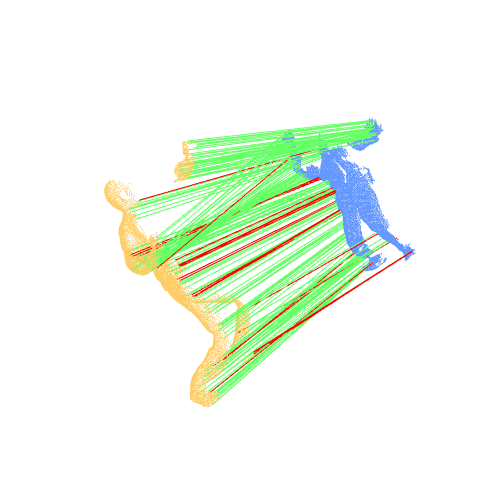} &
		\includegraphics[width=0.094\textwidth,height=1.40cm]
		{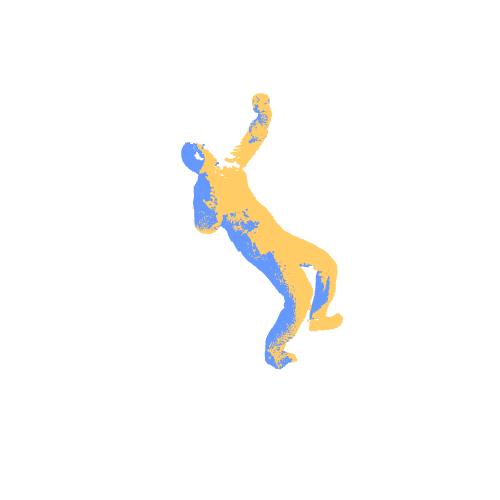} \\
		
		\includegraphics[width=0.094\textwidth,height=1.40cm]
		{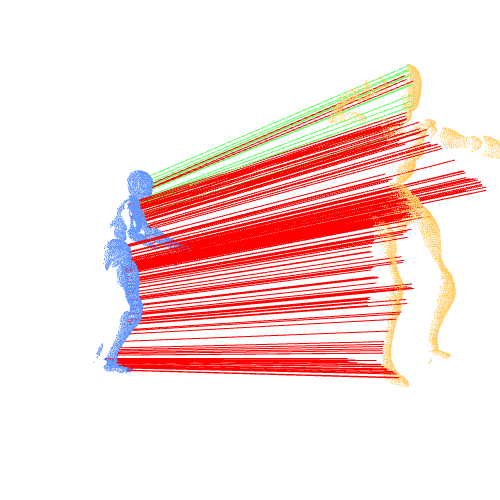} &
		\includegraphics[width=0.094\textwidth,height=1.40cm]
		{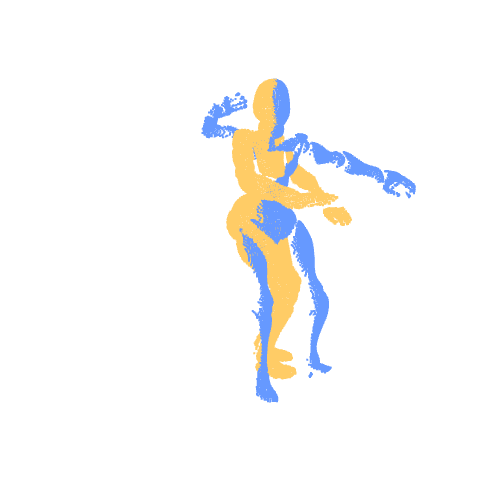} &
		\includegraphics[width=0.094\textwidth,height=1.40cm]
		{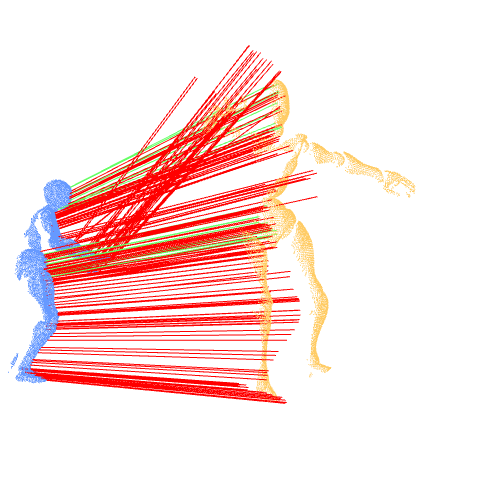} &
		\includegraphics[width=0.094\textwidth,height=1.40cm]
		{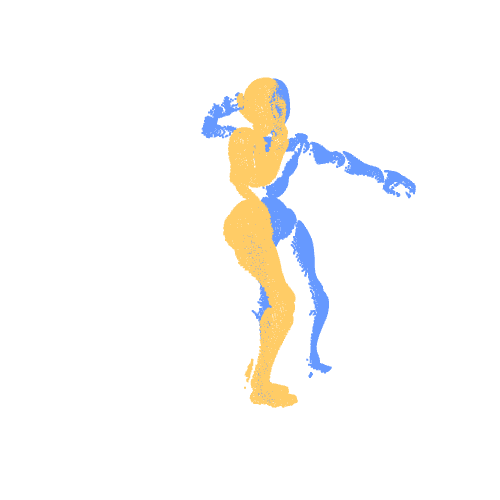} &
		\includegraphics[width=0.094\textwidth,height=1.40cm]
		{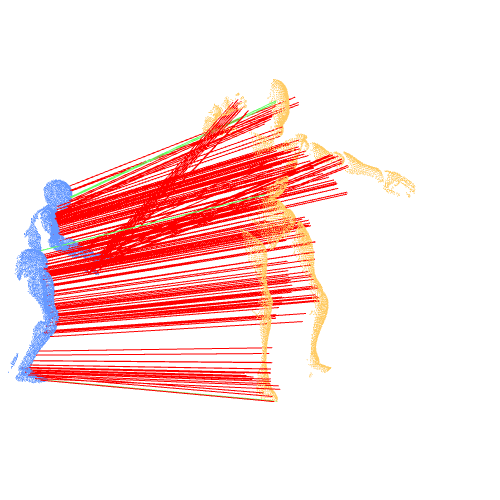} &
		\includegraphics[width=0.094\textwidth,height=1.40cm]
		{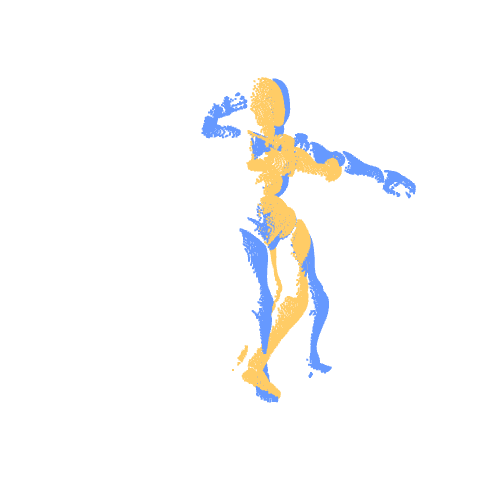} &
		\includegraphics[width=0.094\textwidth,height=1.40cm]
		{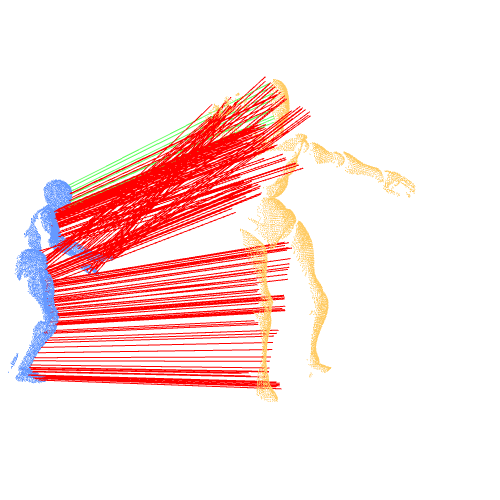} &
		\includegraphics[width=0.094\textwidth,height=1.40cm]
		{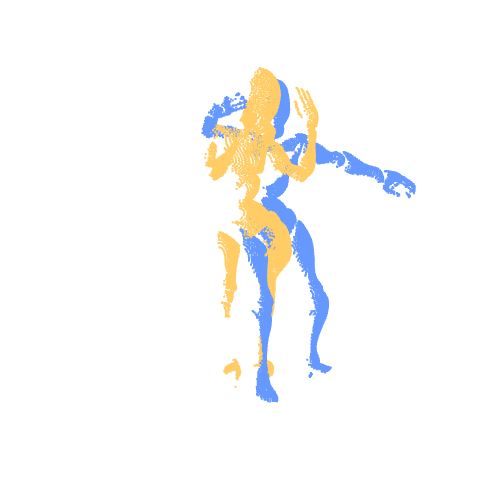} &
		\includegraphics[width=0.094\textwidth,height=1.40cm]
		{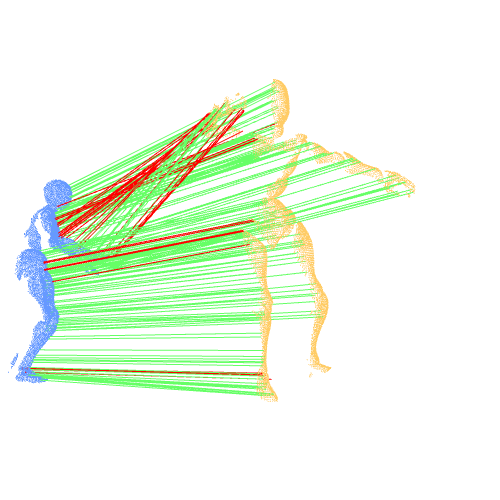} &
		\includegraphics[width=0.094\textwidth,height=1.40cm]
		{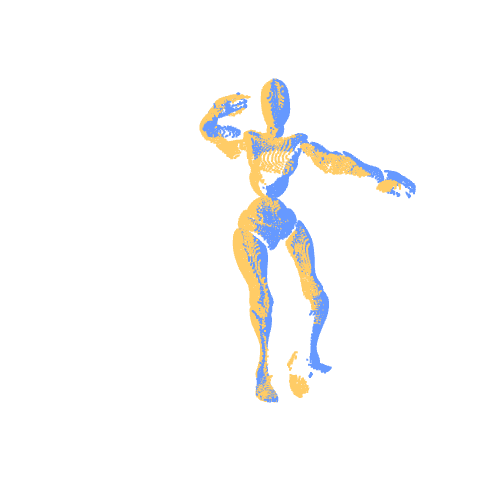} \\
		
		\includegraphics[width=0.094\textwidth,height=1.40cm]
		{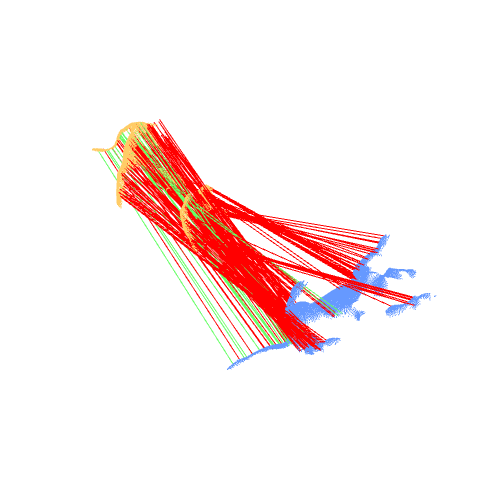} &
		\includegraphics[width=0.094\textwidth,height=1.40cm]
		{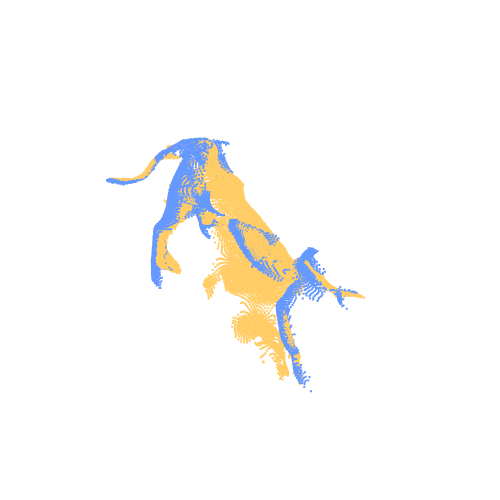} &
		\includegraphics[width=0.094\textwidth,height=1.40cm]
		{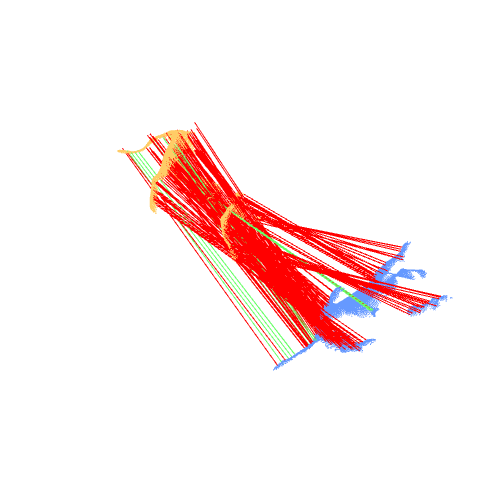} &
		\includegraphics[width=0.094\textwidth,height=1.40cm]
		{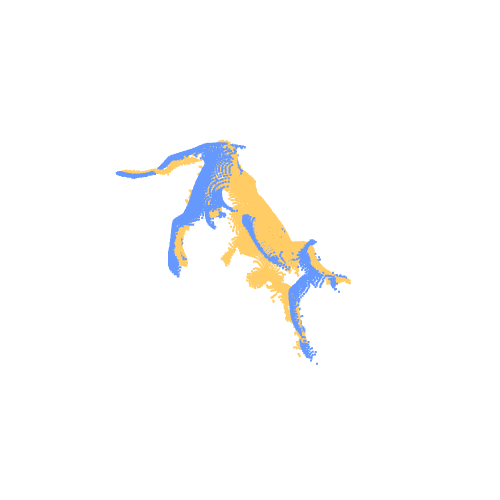} &
		\includegraphics[width=0.094\textwidth,height=1.40cm]
		{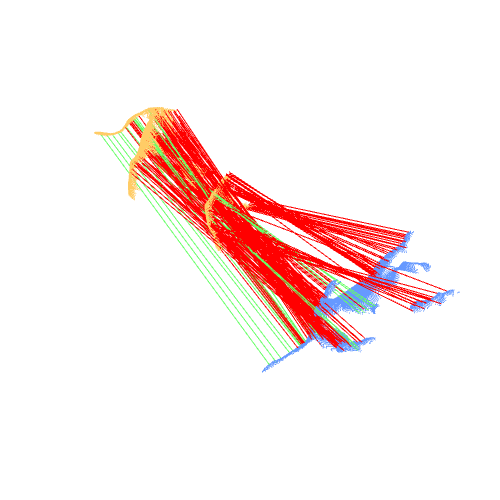} &
		\includegraphics[width=0.094\textwidth,height=1.40cm]
		{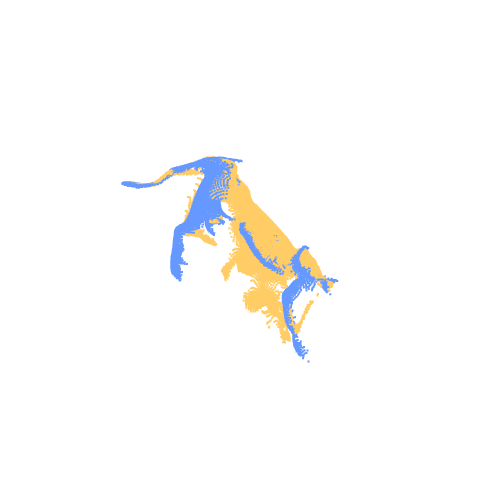} &
		\includegraphics[width=0.094\textwidth,height=1.40cm]
		{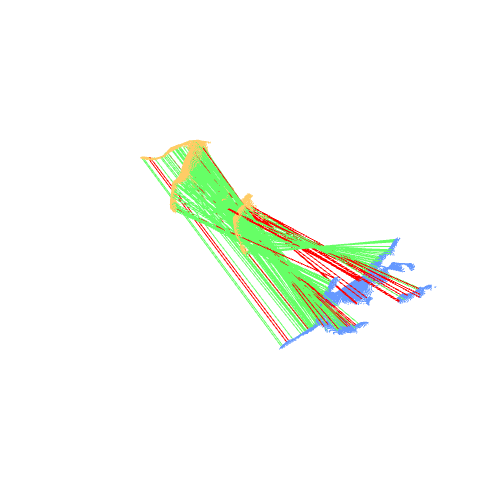} &
		\includegraphics[width=0.094\textwidth,height=1.40cm]
		{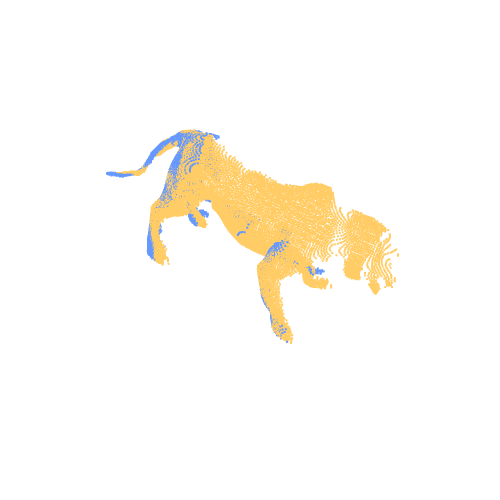} &
		\includegraphics[width=0.094\textwidth,height=1.40cm]
		{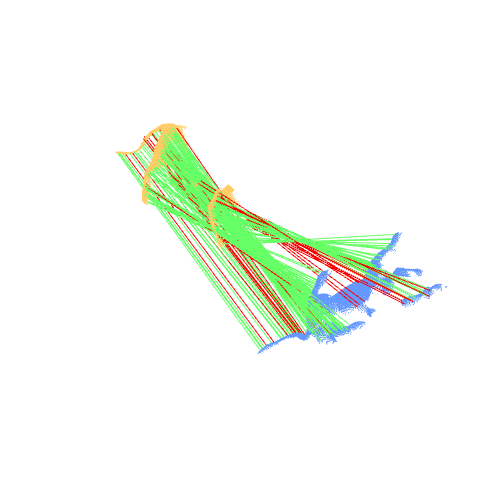} &
		\includegraphics[width=0.094\textwidth,height=1.40cm]
		{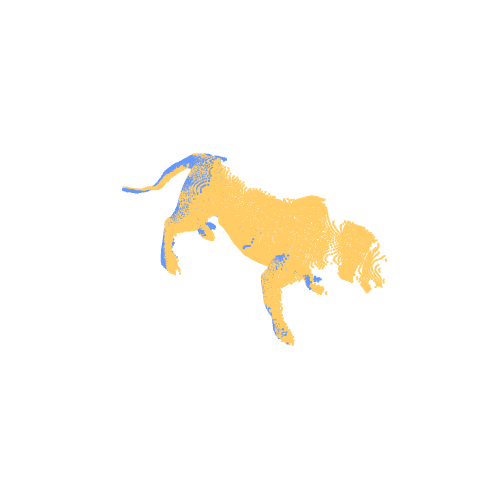} \\
		
		\includegraphics[width=0.094\textwidth,height=1.40cm]
		{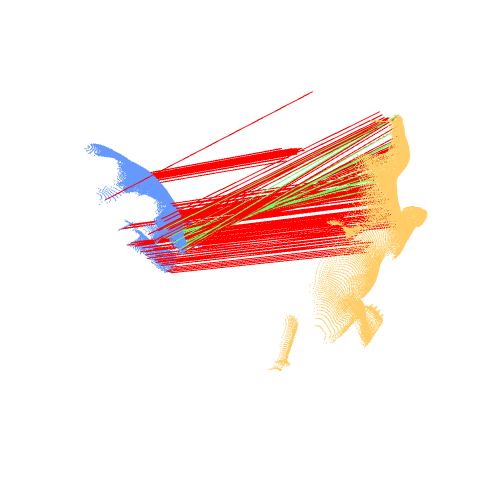} &
		\includegraphics[width=0.094\textwidth,height=1.40cm]
		{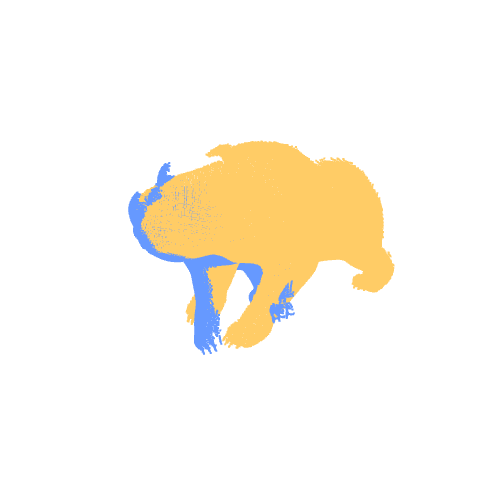} &
		\includegraphics[width=0.094\textwidth,height=1.40cm]
		{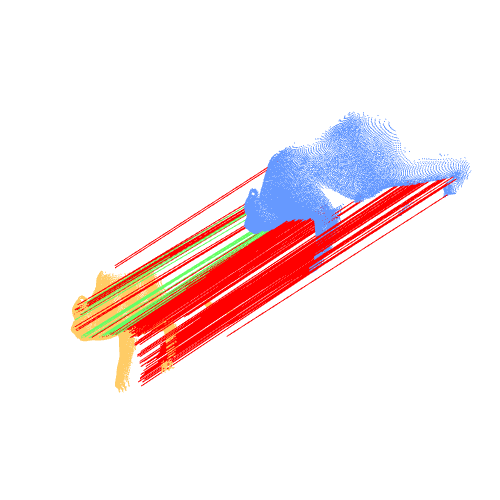} &
		\includegraphics[width=0.094\textwidth,height=1.40cm]
		{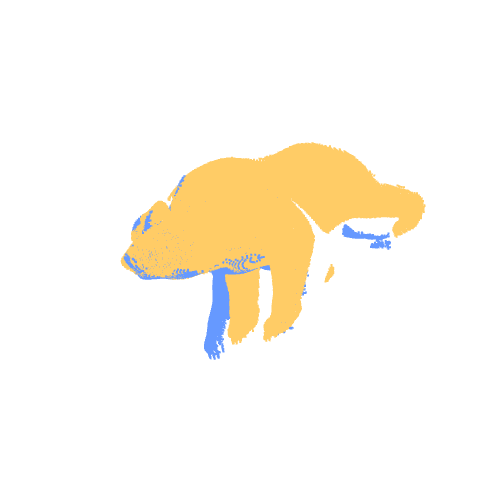} &
		\includegraphics[width=0.094\textwidth,height=1.40cm]
		{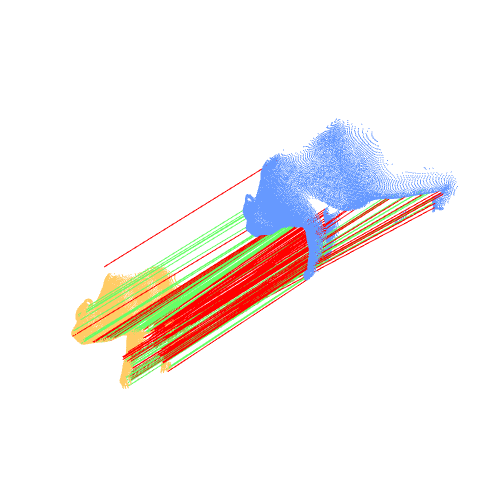} &
		\includegraphics[width=0.094\textwidth,height=1.40cm]
		{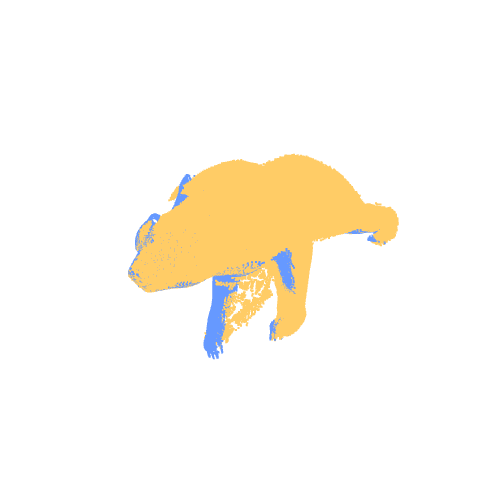} &
		\includegraphics[width=0.094\textwidth,height=1.40cm]
		{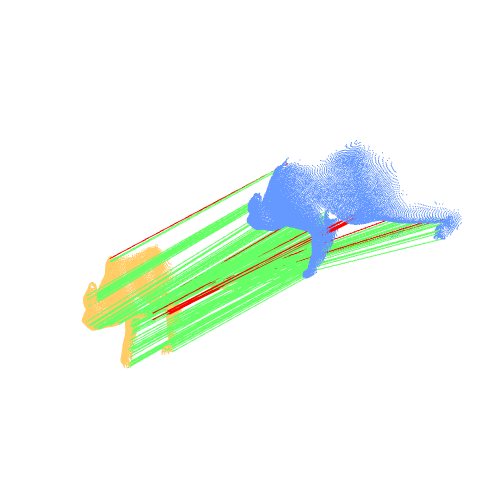} &
		\includegraphics[width=0.094\textwidth,height=1.40cm]
		{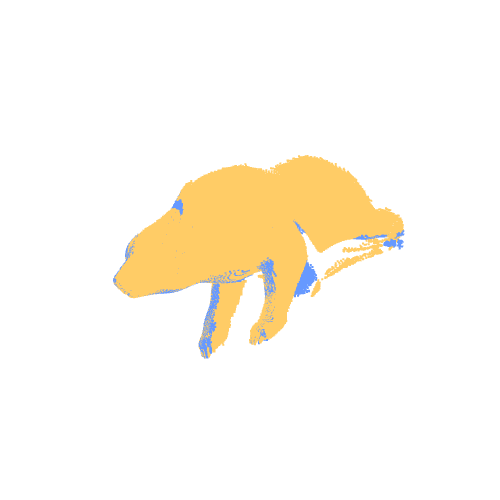} &
		\includegraphics[width=0.094\textwidth,height=1.40cm]
		{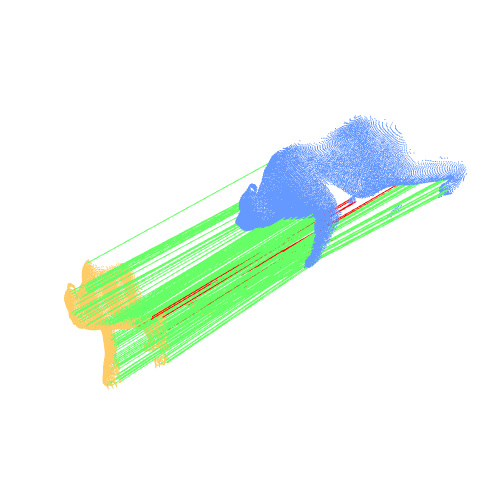} &
		\includegraphics[width=0.094\textwidth,height=1.40cm]
		{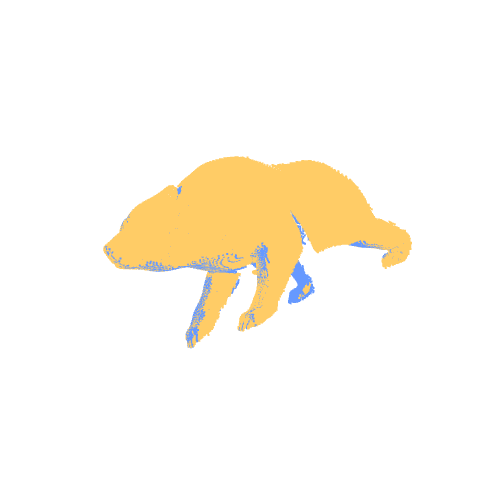}
	\end{tabular}
	
	\caption{
		Qualitative non-rigid registration comparisons on
		4DMatch/4DLoMatch. For each method, the left image visualizes
		predicted correspondences and the right image shows the deformation
		estimated by GraphSCNet. Blue and yellow denote the source and target
		point clouds, respectively; green and red lines indicate correct and
		incorrect correspondences. ODE and Stochastic denote our deterministic
		endpoint-CFM and stochastic Brownian-bridge variants, respectively.
		Zoom in for details.
	}
	\label{fig:non_rigid_registration_vis}
\end{figure*}

\begin{table}[t]
	\caption{Non-rigid correspondence on 4DMatch/4DLoMatch (\%). Bold/underline mark best/second-best. SDE drift ODE: lifted source, one update, 2 Stage-II NFE; initial noise retained, subsequent increments omitted.}
	\label{tab:4d_main}
\setlength{\tabcolsep}{3pt}
	\centering
	\resizebox{\linewidth}{!}{
		\begin{tabular}{lcccc}
			\toprule
			& \multicolumn{2}{c}{4DMatch} & \multicolumn{2}{c}{4DLoMatch}\\
			\cmidrule(lr){2-3}\cmidrule(lr){4-5}
			Method & NFMR $\uparrow$ & IR $\uparrow$ & NFMR $\uparrow$ & IR $\uparrow$\\
			\midrule
			PointPWC~\cite{wu2019pointpwc} 
			& 21.60 & 20.00 & 10.00 & 7.20\\
			
			FLOT~\cite{puy2020flot} 
			& 27.10 & 24.90 & 15.20 & 10.70\\
			
			D3Feat~\cite{bai2020d3feat} 
			& 55.50 & 54.70 & 27.40 & 21.50\\
			
			Predator~\cite{huang2021predator} 
			& 56.40 & 60.40 & 32.10 & 27.50\\
			
			Lepard~\cite{li2022lepard} 
			& 83.60 & 82.64 & 66.63 & 55.55\\
			
			GeoTransformer~\cite{qin2022geometric} 
			& 83.20 & 82.20 & 65.40 & 63.60\\
			
			RoITr~\cite{yu2023rotation} 
			& 83.00 & 84.40 & 69.40 & 67.60\\
			
			Diff-Reg~\cite{wu2024diff} 
			& 90.25 & 87.98 & 77.15 & 67.00\\
			
			RoITr-WN~\cite{yanagi2024learning} 
			& 87.2 & 87.3 & 75.3 & 73.3\\
			
			\midrule
			
			Ours$^{\mathrm{ODE}}$
			& 91.47
			& 89.82
			& 78.64
			& 71.78\\
			
			Ours$^{\mathrm{Stochastic}}$ 
			& \underline{92.43}
			& \underline{91.47}
			& \underline{82.41}
			& \underline{78.79}\\

			\midrule
            Ours$^{\mathrm{SDE\ drift\ ODE}}$ & \textbf{93.10} & \textbf{92.25} & \textbf{83.94} & \textbf{80.49}\\
            \bottomrule
	\end{tabular}}
\end{table}

Both variants improve on Diff-Reg in all four matching metrics (Table~\ref{tab:4d_main}). With one forward-drift update and a final endpoint prediction, the SDE model reaches $93.10/92.25$ NFMR/IR on 4DMatch and $83.94/80.49$ on 4DLoMatch. This setting requires 2 Stage-II network evaluations and achieves the highest scores in all four columns.

\begin{table*}
	\caption{Non-rigid registration on 4DMatch-F/4DLoMatch-F. EPE is in meters; AccS/AccR/OR are percentages. GraphSCNet-labeled entries use this deformation solver~\cite{qin2023deep}; others retain their respective pipelines. RGGT's TF denotes pretrained TRELLIS features; the gray row uses this additional prior and is excluded from ranking. Bold/underline mark best/second-best among the remaining rows.}
	\centering
	\resizebox{\linewidth}{!}{
		\begin{tabular}{lcccccccc}
			\toprule
			\multirow{2}{*}{Method}& \multicolumn{4}{c}{4DMatch-F} & \multicolumn{4}{c}{4DLoMatch-F} \\
			\cmidrule(lr){2-5}\cmidrule(lr){6-9}
			& EPE (m)$\downarrow$ & AccS$\uparrow$ & AccR$\uparrow$ & OR$\downarrow$ & EPE (m)$\downarrow$ & AccS$\uparrow$ & AccR$\uparrow$ & OR$\downarrow$ \\
			\midrule
			
			NSEFP~\cite{li2021neural}
			& 0.265 & 8.7 & 18.7 & 65.0
			& 0.495 & 0.4 & 1.6 & 84.8 \\
			
			Nerfies~\cite{park2021nerfies}
			& 0.280 & 12.7 & 25.4 & 58.9
			& 0.498 & 1.1 & 3.0 & 82.2 \\
			
			PointPWC-Net~\cite{wu2020pointpwc}
			& 0.182 & 6.3 & 21.5 & 52.1
			& 0.279 & 1.7 & 8.2 & 55.7 \\
			
			FLOT~\cite{puy2020flot}
			& 0.133 & 7.7 & 27.2 & 40.5
			& 0.210 & 2.7 & 13.1 & 42.5 \\
			
			DGFM~\cite{donati2020deep}
			& 0.152 & 12.3 & 32.6 & 37.9
			& 0.148 & 1.9 & 6.5 & 64.6 \\
			
			SyNoRiM~\cite{huang2022multiway}
			& 0.099 & 22.9 & 49.9 & 26.0
			& 0.170 & 10.6 & 30.2 & 31.1 \\
			
			NDP~\cite{li2022non}
			& 0.077 & 61.3 & 74.1 & 17.3
			& 0.177 & 26.6 & 41.1 & 33.8 \\
			\midrule

			RoITr\cite{yu2023rotation} $+$GraphSCNet& 0.056 &59.60 &80.50& 12.50&0.118&32.30&56.70&20.50  \\
			Lepard\cite{li2022lepard}$+$GraphSCNet&0.042&70.10 &83.80& 9.20&  0.102& 40.00& 59.10& 17.50  \\
			GeoTR\cite{qin2022geometric}$+$GraphSCNet& 0.043 &72.10 &84.30& 9.50& 0.119& 41.00& 58.40& 20.60   \\
			
			Diff-Reg\cite{wu2024diff}$+$GraphSCNet& 0.041 &\underline{73.20} &85.80& 8.30& 0.095& 43.80&62.90& 15.50   \\
			Lepard+OAR\cite{zhao2025occlusion}  &0.059 &59.32 &74.33 &16.41 &0.251 &27.25&42.01&45.04\\
			RGGT\cite{zhengrggt} (w/o TF)& 0.039&62.10 &79.60&11.3& --&-- &--&--    \\
		\color{gray}{RGGT\cite{zhengrggt} (w TF)}
		& {\color{gray}0.031}
		& {\color{gray}73.9}
		& {\color{gray}89.7}
		& {\color{gray}7.1}
		& {\color{gray}0.052}
		& {\color{gray}55.0}
		& {\color{gray}79.0}
		& {\color{gray}6.8} \\
			\midrule
			Ours$^{\mathrm{ODE}}$+GraphSCNet
			& \underline{0.037}
			& 73.1
			& \underline{86.3}
			& \underline{7.9}
			& \underline{0.094}
			& \underline{45.0}
			& \underline{65.0}
			& \underline{14.9} \\
			
			Ours$^{\mathrm{Stochastic}}$+GraphSCNet
			& \textbf{0.036}
			& \textbf{74.1}
			& \textbf{87.0}
			& \textbf{7.4}
			& \textbf{0.089}
			& \textbf{47.2}
			& \textbf{67.2}
			& \textbf{13.8} \\
			\bottomrule
	\end{tabular}}
	\label{non_rigid_registration}

\end{table*}

Table~\ref{non_rigid_registration} reports registration results using GraphSCNet. Relative to Diff-Reg, SDE lowers EPE from $0.041$ to $0.036$ on 4DMatch-F and from $0.095$ to $0.089$ on 4DLoMatch-F, with higher AccS/AccR and lower OR on both. ODE also lowers EPE on both splits. Figure~\ref{fig:non_rigid_registration_vis} illustrates matching and alignment around limbs and extremities.

\subsection{Zero-Shot Generalization on CAPE}

Following RGGT~\cite{zhengrggt}, we expand 209 four-frame MPC-CAPE samples into 12 ordered pairs each, giving 2,508 test pairs.

\begin{table*}[!t]
	\caption{CAPE generalization (EPE in m; other metrics in \%). Diff-Reg and ours are evaluated on 2,508 pairs. $^*$: target-domain training; --: unverified information. Bold/underline mark best/second-best only in the No group. Stochastic results average three seeds; training and input settings differ across methods.}
	\label{tab:cape_zero_shot}
	\centering
	\resizebox{\textwidth}{!}{
		\begin{tabular}{lllcrrrr}
			\toprule
			Method & Venue & Training data & Target train & EPE (m)$\downarrow$ & AccS$\uparrow$ & AccR$\uparrow$ & OR$\downarrow$\\
			\midrule
			\multicolumn{8}{l}{\textit{Target-domain training}}\\
			SyNoRiM (self-sup., four-view)$^*$~\cite{huang2022multiway} & TPAMI 2022 & CAPE & Yes & 0.030 & 55.5 & 89.1 & 59.1\\
			\midrule
			\multicolumn{8}{l}{\textit{Target-domain training status unverified}}\\
			PointPWC-Net~\cite{wu2020pointpwc} & ECCV 2020 & -- & -- & 0.039 & 17.9 & 35.7 & 85.9\\
			FLOT~\cite{puy2020flot} & ECCV 2020 & -- & -- & 0.049 & 21.2 & 31.1 & 92.2\\
			DGFM~\cite{donati2020deep} & CVPR 2020 & -- & -- & 0.036 & 35.5 & 62.4 & 73.7\\
			Lepard+N-ICP~\cite{li2022lepard} & CVPR 2022 & -- & -- & 0.089 & 23.9 & 44.7 & 78.0\\
			LNDP~\cite{li2022non} & NeurIPS 2022 & -- & -- & 0.045 & 61.3 & 92.2 & 45.7\\
			\midrule
			\multicolumn{8}{l}{\textit{Zero-shot transfer (no target-domain training)}}\\
			GeoTransformer~\cite{qin2022geometric} & CVPR 2022 & ModelNet & No & 0.055 & 20.9 & 50.1 & 75.5\\
			GraphSCNet~\cite{qin2023deep} & CVPR 2023 & 4DMatch & No & 0.059 & 66.7 & 79.4 & 62.1\\
			Diff-Reg~\cite{wu2024diff} + GraphSCNet & ECCV 2024 & 4DMatch & No & 0.018 & 78.4 & 95.1 & 45.4\\			
			RGGT (w TF)~\cite{zhengrggt} & ICML 2026 & ModelNet + 4DMatch & No & 0.023 & 77.7 & 93.2 & \textbf{36.4}\\
			Ours$^{\mathrm{ODE}} +$ GraphSCNet &  & 4DMatch & No & \underline{0.017} & \underline{79.2} & \underline{95.3} & 44.7\\
			Ours$^{\mathrm{Stochastic}} +$ GraphSCNet && 4DMatch & No & \textbf{0.016} & \textbf{80.4} & \textbf{95.7} & \underline{43.4}\\
			\bottomrule
	\end{tabular}}

\end{table*}

With the same GraphSCNet solver, ODE and SDE reduce CAPE EPE from Diff-Reg's $0.018$ to $0.017$ and $0.016$, respectively (Table~\ref{tab:cape_zero_shot}). SDE reaches $80.4/95.7\%$ AccS/AccR. It also has the lowest EPE and OR and the highest AccS in each of the three deformation groups.

SyNoRiM uses target-domain self-supervision and four-view input, and RGGT adds training data and TRELLIS features. Our SDE model leads the zero-shot group in EPE, AccS, and AccR; RGGT has the lowest OR.

\subsection{Zero-Shot Generalization on DeepDeform}

We evaluate 1,221 directed MPC-DD/DeepDeform pairs after near-rigid filtering. We fit a rigid transform on flow-visible points and exclude pairs with mean residual below 1\,cm over the full source cloud. The published GraphSCNet split contains 1,011 pairs. Reference rows are grouped by training and evaluation protocol.

\begin{table*}[!t]
	\caption{DeepDeform generalization (EPE in m; other metrics in \%). $^\dagger$ marks cited results under different protocols; all other rows are evaluated on 1,221 pairs after 1\,cm near-rigid filtering. $^*$ denotes target-domain training or fine-tuning. Bold/underline mark best/second-best only in the No group. Stochastic results average three seeds.}
	\label{tab:deepdeform_zero_shot}
	\centering
	\resizebox{\linewidth}{!}{
		\begin{tabular}{llccrrrr}
			\toprule
			Method & Venue & Training data & Target train & EPE (m)$\downarrow$ & AccS$\uparrow$ & AccR$\uparrow$ & OR$\downarrow$\\
			\midrule
			\multicolumn{8}{l}{\textit{Target-domain training}}\\
			SyNoRiM$^*$~\cite{huang2022multiway} & TPAMI 2022 & MPC-DD & Yes & 0.0268 & 72.9 & 91.8 & 12.8\\
			AniSym-Net$^{*\dagger}$~\cite{wang2025anisym} & TPAMI 2025 & \shortstack{MPI-FAUST\\$\to$ DeepDeform} & Yes & 0.0266 & -- & 89.4 & --\\
			\midrule
			\multicolumn{8}{l}{\textit{Zero-shot transfer (no target-domain training)}}\\
			GeoTransformer~\cite{qin2022geometric}$+$GraphSCNet$^\dagger$ & CVPR 2022 & 4DMatch & No & 0.134 & 24.1 & 44.2 & 59.1\\
			Lepard~\cite{li2022lepard} + GraphSCNet & CVPR 2022 & 4DMatch & No & 0.1501 & 17.1 & 35.1 & 64.0\\
			Diff-Reg~\cite{wu2024diff} + GraphSCNet & ECCV 2024 & 4DMatch & No & 0.0748 & 43.6 & 66.1 & 36.0\\
			Ours$^{\mathrm{ODE}}$ + GraphSCNet &  & 4DMatch & No & \underline{0.0697} & \underline{46.4} & \underline{69.4} & \underline{33.8}\\
			Ours$^{\mathrm{Stochastic}}$ + GraphSCNet &  & 4DMatch & No & \textbf{0.0620} & \textbf{53.2} & \textbf{74.9} & \textbf{28.3}\\
			\bottomrule
	\end{tabular}}

\end{table*}

SDE has the best scores among the matched zero-shot methods on all four metrics (Table~\ref{tab:deepdeform_zero_shot}). Relative to Diff-Reg, EPE decreases from $0.0748$ to $0.0620$ ($17.1\%$), AccS/AccR increase by $9.6/8.8$ points, and OR decreases by $7.7$ points. ODE also improves all four metrics.

\begin{table}[t]
 \caption{Training and inference-source controls for the stochastic model on 4DLoMatch (\%). Results average three sampling seeds per trained model. The last two rows use the same full-model checkpoint; high-resolution inference uses 20 updates. Coarse-only predictions are lifted for evaluation.}
 \label{tab:hierarchy_main}
 \centering\small
 \setlength{\tabcolsep}{3pt}
 \resizebox{\linewidth}{!}{\begin{tabular}{llrr}
  \toprule
  Training & Inference source / output & NFMR$\uparrow$ & IR$\uparrow$\\
  \midrule
  Coarse-only & Lifted coarse output & 34.04 & 18.03\\
  High-res.-only & $\mathcal U(0,1)$ source & 80.27 & 75.98\\
  Full two-stage & Lifted Stage-I source & 82.41 & 78.79\\
  Full two-stage (same weights) & $\mathcal U(0,1)$ source & 82.43 & 78.92\\
  \bottomrule
 \end{tabular}}
\end{table}

\begin{table}[t]
 \caption{Inference controls on one SDE-trained checkpoint, 4DLoMatch (\%). Within each source, all modes share the initial perturbation. Forward-drift and probability-flow updates omit subsequent random increments and include a final endpoint prediction. Bold marks the higher scores at 2 NFE within each source.}
 \label{tab:sde_inference_main}
 \centering\small
 \setlength{\tabcolsep}{3pt}
 \resizebox{\linewidth}{!}{\begin{tabular}{llrrr}
  \toprule
  Source & Inference mode & Stage-II NFE & NFMR$\uparrow$ & IR$\uparrow$\\
  \midrule
  Lifted & Direct endpoint & 1 & 79.28 & 76.69\\
  Lifted & Forward drift & 2 & \textbf{83.94} & \textbf{80.49}\\
  Lifted & Probability flow & 2 & 74.06 & 72.56\\
  \midrule
  Uniform & Direct endpoint & 1 & 75.12 & 72.79\\
  Uniform & Forward drift & 2 & \textbf{82.82} & \textbf{79.61}\\
  Uniform & Probability flow & 2 & 74.85 & 73.02\\
  \bottomrule
 \end{tabular}}
\end{table}

\subsection{Ablations and Solver Budgets}
\label{sec:ablation_main}

\paragraph{Multiresolution design and initialization.}
The full SDE design gains $2.14/2.81$ NFMR/IR points over high-resolution-only (Table~\ref{tab:hierarchy_main}). At twenty updates, lifted and uniform sources reach $82.41/78.79$ and $82.43/78.92$. Lifting benefits one-update ODE/SDE inference; longer refinement performs well from either source.

\paragraph{SDE forward-drift refinement without random increments.}
Here we retain the initial perturbation and learned noise correction but omit subsequent Brownian increments. One forward-drift update followed by endpoint prediction raises lifted-source NFMR/IR from $79.28/76.69$ to $83.94/80.49$, using 2 NFE; direct endpoint prediction uses 1 (Table~\ref{tab:sde_inference_main}). At 2 NFE, forward drift exceeds probability flow by $9.88/7.93$ points. Forward drift has higher accuracy at every tested budget.

\paragraph{Standard ODE and stochastic SDE sampling budgets.}
ODE refinement raises NFMR/IR from $75.76/70.11$ at one step to $78.46/71.71$ at ten. Standard SDE sampling, with Brownian increments retained at each update, peaks at two updates with $83.16/79.62$. These settings take $355$ and $64$\,ms in Stage~II, respectively; twenty updates take $740$ and $639$\,ms.

\paragraph{Candidate coverage.}
Full-candidate refinement recovers correct matches excluded by coarse selection. Top-256 retains $34.23\%$ of supervised pairs; unrestricted inference recovers $49.30\%$ of the excluded pairs and gains $33.10$ NFMR points. This fixed-model masking experiment uses dense computation; the matching state requires $O(N_{\mathrm{hr}}M_{\mathrm{hr}})$ memory.

\section{Conclusion}
\label{sec:conclusion}

We presented \paper, a two-stage model for non-rigid point cloud registration. It lifts a coarse diffusion prediction to a complete high-resolution matching matrix, then refines the matrix with geometry-conditioned ODE or SDE updates. Experiments show improvements in matching, registration, and transfer to unseen datasets. Endpoint-CFM benefits from several updates. For the SDE model, one forward-drift update without subsequent random increments, followed by endpoint prediction, gives accurate results with two network evaluations.

\FloatBarrier
{\small\bibliography{main}}
\bibliographystyle{ieeenat_fullname}

\clearpage
\appendix
\section*{Appendix}
\section{Technical Background}
\label{sec:background}

We summarize the path definitions used by our solvers. Let $z=(\X_0,\Y)\sim\pi_c$ denote paired matrices under geometric condition $c$, and let $t\in[0,1]$.

\subsection{Flow Matching}
\label{sec:bg_fm}

Conditional Flow Matching learns a velocity from an analytic conditional path without simulating training trajectories~\cite{lipman2023flowmatching}. For conditional velocity $u_t$, its squared objective is
\begin{equation}
 \mathcal{L}_{\mathrm{CFM}}
 =\E_{t,z,\X_t}\!\left[\|v_\phi(\X_t,t,c)-u_t(\X_t\mid z,c)\|_F^2\right].
 \label{eq:canonical_fm_loss}
\end{equation}
The population optimum of the squared loss is $v_t^\star(\X,c)=\E[u_t\mid\X_t=\X,c]$. For the straight path $\X_t=(1-t)\X_0+t\Y$, $u_t=\Y-\X_0$~\cite{liu2022flow}. Our endpoint-CFM model defines the velocity as $v_\phi=(\widehat{\Y}_t-\X_t)/(1-t)$ and trains the endpoint predictor with focal correspondence loss. Section~\ref{sec:additional_derivations} relates squared velocity error to endpoint error. Section~\ref{sec:training} gives the training loss.

\subsection{Schr\"odinger Bridges}
\label{sec:bg_sb}

Given reference diffusion $\mathbb P^\sigma$ and endpoint marginals $\mu_0,\mu_1$, the classical Schr\"odinger bridge minimizes the Kullback--Leibler divergence between path distributions~\cite{leonard2014survey,debortoli2021diffusion}:
\begin{equation}
 \mathbb Q^\star=\arg\min_{\mathbb Q}\KL(\mathbb Q\|\mathbb P^\sigma),
 \qquad \mathbb Q_0=\mu_0,\quad\mathbb Q_1=\mu_1.
 \label{eq:classic_sb_background}
\end{equation}
Our paired Brownian construction fixes the coupling through supervised source--target pairs. Conditioned on an endpoint pair $z=(\X_0,\Y)$, the Brownian path has noise scale $\rho(t)=\sigma_Bs_t$, where $s_t=\sqrt{t(1-t)}$~\cite{tong2024simulation}. Writing $q_t$ for this endpoint-conditioned density, its score and drift satisfy
\begin{equation}
 \nabla_{\X_t}\log q_t=-\frac{\bm\epsilon}{\sigma_Bs_t},
 \qquad b_t=u_t+\frac{\sigma_B^2}{2}\nabla_{\X_t}\log q_t.
 \label{eq:sde_pf_relation}
\end{equation}
These identities hold for the given endpoint pair and guide the design of the Brownian sampler. We train the endpoint and noise predictors with the losses in Sec.~\ref{sec:training}.

\subsection{A Shared Path Family}
\label{sec:bg_unified}

Both branches use the following path between the source matrix $\X_0$ and target matrix $\Y$:
\begin{equation}
 \X_t=(1-t)\X_0+t\Y+\rho(t)\bm\epsilon,
 \qquad \bm\epsilon\sim\N(\mathbf 0,\I).
 \label{eq:unified_path_reparam}
\end{equation}
Holding the endpoints and noise fixed gives $u_t=\Y-\X_0+\dot\rho(t)\bm\epsilon$. Zero variance gives endpoint-CFM; $\rho(t)=\sigma_B\sqrt{t(1-t)}$ gives the Brownian path. With exact fields, suitable regularity, and the same initial marginal at an interior time, the Brownian SDE and its probability-flow ODE share time marginals. The zero-variance branch follows a different path. Section~\ref{sec:training_objective_scope} describes how the task losses train the two predictors.

\section{Solver and Training Details}

\paragraph{Model and notation.}
\paper estimates correspondences between source points $\{\mathbf p_i\}$ and target points $\{\mathbf q_j\}$. A shared backbone produces coarse ($\mathrm{cr}$) and high-resolution ($\mathrm{hr}$) features. Stage~I denoises a coarse matching matrix; Stage~II refines a high-resolution matrix. The binary matrices $\Y^{(r)}$ provide correspondence supervision at resolution $r$. One-hot parent assignments $\mathbf P_s,\mathbf P_t$ lift coarse scores to
\begin{equation}
 \X_0=\U(\widehat{\Y}^{(\mathrm{cr})})
 =\mathbf P_s\widehat{\Y}^{(\mathrm{cr})}\mathbf P_t^\top.
\end{equation}
Each coarse entry is copied to pairs with the same parents, keeping all candidates. We omit $\mathrm{hr}$ when describing Stage~II. During training, $\X_{0,\mathrm{tr}}$ denotes the source selected by the training schedule in Sec.~\ref{sec:training}. In the training-path formulas, $\X_0$ is shorthand for this source. At inference, it denotes the lifted DDIM estimate unless the experiment replaces the source. The two bridge variants have separate weights. Both use $H_\phi$ to predict the endpoint, and the stochastic model also uses $G_\psi$ to predict noise.

\subsection{Matching-Matrix Endpoint Predictor}
\label{sec:endpoint_predictor_details}

The endpoint predictor $H_{\phi}$ consists of a repositioning transformer and a Sinkhorn matching head. Its input matrix is $\mathbf C_t=\X_t$ for the bounded ODE state and $\mathbf C_t=\clip(\X_t,0,1)$ for the unbounded Brownian state. After invalid entries are masked, three Sinkhorn iterations with a learned dustbin score supply weights for Soft Procrustes. The estimated transform $(\mathbf{R}_t,\bm\tau_t)$ warps source points to $\widetilde{\mathbf p}_{i,t}=\mathbf R_t\mathbf p_i+\bm\tau_t$, which determine the rotary positional encodings. The same $\mathbf{C}_t$ is also added to the final pairwise logits.

The high-resolution backbone features have dimension $d=528$. A two-layer ReLU MLP processes the sinusoidal time embedding, which is added to source and target features. Six geometry-attention blocks follow in the order
\begin{equation}
 [\mathrm{self},\mathrm{cross},\mathrm{self},\mathrm{cross},
   \mathrm{self},\mathrm{cross}],
\end{equation}
with four attention heads and rotary geometric positional encoding in each block. Let $\mathbf h_{s,i}$ and $\mathbf h_{t,j}$ denote the transformer output features after the learned matching projections, and let $\bar{\mathbf h}=\mathbf h/\sqrt d$. The pre-Sinkhorn endpoint logits are
\begin{equation}
 \ell_{t,ij}=\bar{\mathbf h}_{s,i}^{\top}\bar{\mathbf h}_{t,j}
 + C_{t,ij}.
 \label{eq:endpoint_logits}
\end{equation}
After masking invalid pairs, we apply the same three Sinkhorn iterations with a dustbin to obtain
\begin{equation}
 \widehat{\Y}_t
 =H_{\phi}(\X_t,t,c)
 =\operatorname{Crop}\!\left(
 \operatorname{Sinkhorn}_{3}(\boldsymbol\ell_t;\alpha_{\mathrm{bin}})
 \right),
 \label{eq:endpoint_predictor_impl}
\end{equation}
Here, $\alpha_{\mathrm{bin}}$ is learned and $\operatorname{Crop}$ removes the dustbin row and column. Mutual nearest matching with threshold $0.2$ extracts correspondences from $\widehat{\Y}_t$. The ODE and stochastic models have separate weights but use this same architecture. The stochastic model adds a noise head to this endpoint predictor.

\paragraph{Noise head.}
The noise head shares transformer features with the endpoint head. It projects source and target features separately and computes their pairwise inner products. A small time-conditioned network weights the current state and source matrix. We combine these terms with the scaled inner products to predict signed noise. Invalid pairs are masked. Both heads backpropagate through the shared transformer and backbone.

\subsection{Training Objectives and Schedules}
\label{sec:training}
\label{sec:training_main}
\label{sec:experiment_training}

\paragraph{Implementation.}
Both variants use the KPFCN-style hierarchy and preprocessing from Lepard. The outputs of the deepest encoder and the first decoder provide coarse- and high-resolution features, respectively. Stage~I uses deterministic DDIM ($\eta=0$). Unless stated otherwise, both ODE and SDE use 5 Stage-I DDIM steps at the coarse resolution and 20 Stage-II bridge steps at the high resolution. The stochastic setting uses $\sigma_B=0.1$ and $\varepsilon=0.02$.

Each run jointly trains Stage~I and one Stage-II model. Stage~I predicts a clean endpoint from a noisy coarse target. Its detached prediction is lifted to initialize Stage~II. Detachment blocks gradients through the source construction, including DDIM sampling, while both correspondence losses update the shared backbone.

\paragraph{Training supervision schedule.}
We begin training with a detached one-pass coarse prediction. We then gradually increase the probability of using five-step DDIM predictions until every sample uses them. In the final phase, we sample uniformly among a random coarse matrix and detached DDIM predictions with one to five steps. We lift each selected source to high resolution.

During the first three phases, the lifted source is mixed with the high-resolution target:
\begin{equation}
 \X_{0,\mathrm{tr}}=\widetilde{\X}_0
 =\alpha_{\mathrm{mix}}\Y^{(\mathrm{hr})}+(1-\alpha_{\mathrm{mix}})\X_0,
 \label{eq:gt_mix}
\end{equation}
The mixing weight $\alpha_{\mathrm{mix}}$ decreases linearly from $0.3$ to zero and remains zero in the final phase. We keep endpoint supervision and the losses for each branch throughout training. Inference uses the configured DDIM budget and no ground truth.

\paragraph{Losses.}
Focal correspondence losses supervise the coarse and high-resolution endpoint predictions against their binary target matrices. Let $C_{ij}$ be the predicted confidence clipped to $[10^{-6},1-10^{-6}]$. For positive entries $\mathcal P=\{(i,j):Y_{ij}=1\}$ and negative entries $\mathcal N=\{(i,j):Y_{ij}=0\}$ in the batched target tensor, we use
\begin{equation}
\begin{aligned}
 \mathcal L_{\mathrm{corr}}(\mathbf C,\mathbf Y)
 ={}&-\frac{\alpha}{|\mathcal P|}\sum_{(i,j)\in\mathcal P}(1-C_{ij})^{\gamma}\log C_{ij}\\
 &-\frac{\alpha}{|\mathcal N|}\sum_{(i,j)\in\mathcal N}C_{ij}^{\gamma}\log(1-C_{ij}),
\end{aligned}
 \label{eq:focal_corr_loss}
\end{equation}
where $\alpha=0.25$ and $\gamma=2$, and each class mean is taken over the batch. The Sinkhorn loss excludes the dustbin; its negative-class denominator includes zero-target padding entries. Empty positive or negative sets are omitted. The motion term compares predicted and ground-truth displacements on overlapping source points. Let $\mathcal O$ be this source set, $\widehat{\mathbf d}_i=\mathbf R_t\mathbf p_i+\bm\tau_t-\mathbf p_i$, and $\mathbf d_i^{\star}$ the ground-truth displacement after applying the benchmark alignment to the annotated scene flow. Then
\begin{equation}
 \mathcal L_{\mathrm{mot}}
 =\frac{1}{|\mathcal O|}\sum_{i\in\mathcal O}
 \|\widehat{\mathbf d}_i-\mathbf d_i^{\star}\|_1.
 \label{eq:motion_loss}
\end{equation}
Following the matching-backbone implementation, we enable this term once the current batch matching recall exceeds $0.01$. The shared objective is
\begin{equation}
\begin{aligned}
	\mathcal{L}_{\mathrm{shared}}={}&
	\mathcal{L}_{\mathrm{corr}}(\widehat{\Y}^{(\mathrm{cr})},\Y^{(\mathrm{cr})})
	\\&+\mathcal{L}_{\mathrm{corr}}(\widehat{\Y}_t,\Y^{(\mathrm{hr})})
	+\lambda_{\mathrm{mot}}\mathcal{L}_{\mathrm{mot}}.
\end{aligned}
	\label{eq:shared_loss}
\end{equation}
For the stochastic model, let $\mathcal{P}$ and $\mathcal{N}$ denote valid positive and negative entries. Their noise errors are averaged separately to balance the sparse positives against the negative background:
\begin{equation}
	\mathcal{L}_{\epsilon}
	=\frac{w_+\mathbf{1}_{|\mathcal P|>0}\,\mathrm{MSE}_{\mathcal P}
		+w_-\mathbf{1}_{|\mathcal N|>0}\,\mathrm{MSE}_{\mathcal N}}
	{w_+\mathbf{1}_{|\mathcal P|>0}+w_-\mathbf{1}_{|\mathcal N|>0}},
	\label{eq:balanced_noise_loss}
\end{equation}
Here, $\mathrm{MSE}_{\mathcal P}$ and $\mathrm{MSE}_{\mathcal N}$ are the mean squared errors between predicted and injected noise over the valid positive and negative entries, respectively. We use $w_+=w_-=0.5$; the indicators omit empty classes from the weighted average. Hard-negative noise error is logged separately and does not enter the objective. The total loss is
\begin{equation}
	\mathcal{L}=\mathcal{L}_{\mathrm{shared}}
	+\mathbb{I}_{\mathrm{SDE}}\lambda_{\epsilon}\mathcal{L}_{\epsilon}.
	\label{eq:unified_loss}
\end{equation}

Here, $\mathbb{I}_{\mathrm{SDE}}$ equals one for the stochastic branch and zero for endpoint-CFM. On 4DMatch, $\lambda_{\mathrm{mot}}=0.1$ and $\lambda_{\epsilon}=1$.

\paragraph{Deterministic endpoint-CFM bridge.}
For endpoint-CFM, we sample $t\sim\mathcal{U}(0,1)$ and interpolate between the curriculum source and ground truth. Focal correspondence loss and the optional motion loss supervise the endpoint; Eq.~\plaineqref{eq:app_endpoint_velocity} supplies the velocity. The training objective is $\mathcal{L}_{\mathrm{shared}}$.

\paragraph{Stochastic paired Brownian bridge.}
For the stochastic model, we sample $t\sim\mathcal{U}(\varepsilon,1-\varepsilon)$ and $\bm\epsilon\sim\mathcal{N}(\mathbf{0},\mathbf{I})$, then construct $\X_t$ with Eq.~\plaineqref{eq:sb_training_path}. The endpoint head uses the same focal correspondence and motion losses as endpoint-CFM. The noise head receives the additional loss $\lambda_{\epsilon}\mathcal{L}_{\epsilon}$, with $\lambda_{\epsilon}=1$.

\paragraph{Task-weighted field learning.}
\label{sec:training_objective_scope}
Positive matches are sparse in high-resolution matrices, so averaging errors over all entries gives most of the weight to negatives. Focal loss emphasizes misclassified correspondences, and class-balanced noise regression assigns equal total weight to positive and negative entries. The identities in Sec.~\ref{sec:additional_derivations} map the endpoint and noise predictions to velocity and drift.

\paragraph{Optimization details.}
Both variants are trained for 60 epochs with a batch size of 2, an SGD learning rate of $0.015$, momentum of $0.93$, weight decay of $10^{-6}$, exponential decay of $0.95$, and gradient clipping at 5. The Stage-I curriculum uses at most five DDIM steps. The stochastic configuration uses $w_+=w_-=0.5$ and $\lambda_{\epsilon}=1$. The motion-loss weight is $0.1$ for 4DMatch and zero for 3DMatch.

\subsection{Solver Parameterization and Sampling Details}
\label{sec:solver_details}

\paragraph{Stage-I DDIM sampling.}
Let $\bar\alpha_k=\prod_{j=1}^{k}(1-\beta_j)$ be the cumulative signal coefficient, where $\beta_j$ is the forward diffusion noise variance at step $j$. The coarse training state is $\X_k^{(\mathrm{cr})}=\sqrt{\bar\alpha_k}\Y^{(\mathrm{cr})}+\sqrt{1-\bar\alpha_k}\bm\epsilon$, with standard Gaussian noise $\bm\epsilon$. A time-conditioned matching transformer predicts the clean matrix $\widehat{\Y}_{0,k}^{(\mathrm{cr})}$ from this state and coarse point features. At inference, $\X_T^{(\mathrm{cr})}\sim\N(\mathbf{0},\I)$ is updated with deterministic DDIM~\cite{song2020denoising}. From the clean prediction, the estimated noise is
\begin{equation}
	\widehat{\bm\epsilon}_k=
	\frac{\X_k^{(\mathrm{cr})}-\sqrt{\bar\alpha_k}\widehat{\Y}_{0,k}^{(\mathrm{cr})}}
	{\sqrt{1-\bar\alpha_k}},
\end{equation}
and the $\eta=0$ update to $k'<k$ is
\begin{equation}
	\X_{k'}^{(\mathrm{cr})}=\sqrt{\bar\alpha_{k'}}\widehat{\Y}_{0,k}^{(\mathrm{cr})}
	+\sqrt{1-\bar\alpha_{k'}}\widehat{\bm\epsilon}_k.
	\label{eq:l1_ddim}
\end{equation}
The final prediction $\widehat{\Y}_{\mathrm{DDIM}}^{(\mathrm{cr})}$ is lifted to form the Stage-II source.

\paragraph{Deterministic path and Euler integration.}
The deterministic model sets $\rho_{\mathrm{ODE}}(t)=0$ in Eq.~\plaineqref{eq:unified_path_reparam}, giving the straight conditional path
\begin{equation}
	\X_t=(1-t)\X_0+t\Y^{(\mathrm{hr})},
	\label{eq:fm_train_path}
\end{equation}
We use $K_{\mathrm{II}}=20$ explicit Euler steps. With $t_k=k/K_{\mathrm{II}}$ and $\Delta t=1/K_{\mathrm{II}}$, we evaluate
\begin{equation}
\begin{aligned}
	\widehat{\Y}_k&=H_{\phi}(\X_k,t_k,\text{geometry/features}),\\
	\X_{k+1}&=\clip\!\left(
	\X_k+\Delta t\,
	\frac{\widehat{\Y}_k-\X_k}{1-t_k},0,1\right).
\end{aligned}
	\label{eq:fm_euler}
\end{equation}
We evaluate the field at $t_0,\ldots,t_{K_{\mathrm{II}}-1}$, before the singular endpoint. The last update equals the clipped endpoint prediction (Sec.~\ref{sec:additional_derivations}). Earlier updates change the matrix used for geometric conditioning.

\paragraph{Brownian path, noise head, and drift construction.}
The stochastic model sets $\rho_{\mathrm{SDE}}(t)=\sigma_B\sqrt{t(1-t)}$, giving the paired Brownian-bridge path
\begin{equation}
	\X_t=(1-t)\X_0+t\Y^{(\mathrm{hr})}
	+\sigma_B s_t\bm\epsilon,
	\quad s_t=\sqrt{t(1-t)},
	\label{eq:sb_training_path}
\end{equation}
where $\bm\epsilon\sim\N(\mathbf{0},\I)$. A noise head predicts the injected unit noise from the current state, lifted source, time, and shared transformer features:
\begin{equation}
 \widehat{\bm\epsilon}_t
 =G_{\psi}(\X_t,\X_0,t,\mathbf f_s,\mathbf f_t).
 \label{eq:noise_prediction}
\end{equation}
Here, $\mathbf f_s$ and $\mathbf f_t$ are the source and target features shared with the endpoint head. The noise head uses the unclipped matching states. Section~\ref{sec:endpoint_predictor_details} summarizes its design.

Endpoint and noise estimates define a conditional velocity and scaled score,
\begin{align}
	v_{\phi}(\X_t,t)
	 &=\widehat{\Y}_t-\X_0
	+\sigma_B\frac{1-2t}{2s_t}\widehat{\bm\epsilon}_t,
	\label{eq:sb_velocity}\\
	\widetilde{s}_{\phi}(\X_t,t)
	&=-\widehat{\bm\epsilon}_t/s_t.
	\label{eq:sb_score}
\end{align}
The scaled score and velocity give
\begin{equation}
	b_{\phi}=v_{\phi}+\frac{\sigma_B}{2}\widetilde{s}_{\phi}.
	\label{eq:sb_drift}
\end{equation}
Substitution gives the sampling dynamics
\begin{align}
 b_\phi(\X_t,t)&=\widehat{\Y}_t-\X_0-\sigma_B\frac{t}{s_t}\widehat{\bm\epsilon}_t,
 \label{eq:app_sb_drift_explicit}\\
 \mathrm d\X_t&=b_\phi(\X_t,t)\,\mathrm dt+\sigma_B\,\mathrm d\mathbf W_t,
 \label{eq:app_sb_forward_sde}
\end{align}
where $\mathbf W_t$ is a standard matrix-valued Wiener process.

\paragraph{Stochastic initialization and integration.}
For sampling, we use $\X_0=\U(\widehat{\Y}_{\mathrm{DDIM}}^{(\mathrm{cr})})$ and initialize the truncated bridge by
\begin{equation}
	\X_{\varepsilon}=\X_0+\sigma_Bs_{\varepsilon}\mathbf{z}_0,
	\qquad \mathbf{z}_0\sim\N(\mathbf{0},\I),
	\label{eq:sb_initialization}
\end{equation}
Conditioned on both endpoints, the training state at $\varepsilon$ has mean $\X_0+\varepsilon(\Y-\X_0)$ and covariance $\sigma_B^2\varepsilon(1-\varepsilon)\I$. Equation~\plaineqref{eq:sb_initialization} keeps this covariance and replaces the mean by $\X_0$, since the target is unknown at inference. The omitted mean shift is $\varepsilon(\Y-\X_0)$, with Frobenius norm $\varepsilon\|\Y-\X_0\|_F$. We then apply Euler--Maruyama on a uniform grid from $\varepsilon$ to $1-\varepsilon$:
\begin{equation}
	\X_{t+\Delta t}=\X_t+\Delta t\,b_{\phi}(\X_t,t)
	+\sigma_B\sqrt{\Delta t}\,\mathbf{z}_t,
	\quad \mathbf{z}_t\sim\N(\mathbf{0},\I).
	\label{eq:euler_maruyama}
\end{equation}
Noise is zero on padded entries; valid states remain unbounded. After 20 steps, we evaluate the endpoint at $1-\varepsilon$ and clip it to obtain the confidence matrix. The corresponding probability-flow ODE uses $v_{\phi}$ in place of $b_{\phi}$ and removes the stochastic increments. Section~\ref{sec:additional_derivations} gives the initialization error introduced by truncation.

\subsection{Additional Derivations for the Two-Stage Solver}
\label{sec:additional_derivations}

\paragraph{Structure of the lifted source.}
For parent-assignment matrices $\mathbf P_s$ and $\mathbf P_t$, lifting gives $\X_0=\mathbf P_s\widehat{\Y}^{(\mathrm{cr})}\mathbf P_t^{\top}$, hence
\begin{equation}
 \operatorname{rank}(\X_0)\leq\operatorname{rank}(\widehat{\Y}^{(\mathrm{cr})})
 \leq\min(N_{\mathrm{cr}},M_{\mathrm{cr}}).
 \label{eq:rank_bound}
\end{equation}
The lifted matrix is block-constant; high-resolution refinement resolves differences within each parent block.

\paragraph{Orthogonal decomposition of lifting error.}
Let $\Pi$ average a high-resolution matrix within each parent-induced block, and define $\mathcal{P}_{\mathrm{blk}}=\U\Pi$ as the Frobenius-orthogonal projector onto the block-constant subspace. For any compact coarse-resolution matrix $\C$, orthogonality gives
\begin{equation}
\begin{aligned}
	\|\Y^{(\mathrm{hr})}-\U(\C)\|_F^2
	={}&\|(\I-\mathcal{P}_{\mathrm{blk}})\Y^{(\mathrm{hr})}\|_F^2 \\
	&+\|\mathcal{P}_{\mathrm{blk}}\Y^{(\mathrm{hr})}-\U(\C)\|_F^2.
\end{aligned}
	\label{eq:error_decomposition}
\end{equation}
The first term contains within-block details absent from any lifted matrix. The second is the error in the block-constant component.

\paragraph{Endpoint-induced velocity.}
The endpoint prediction defines
\begin{equation}
 v_\phi(\X_t,t)=\frac{\widehat{\Y}_t-\X_t}{1-t}.
 \label{eq:app_endpoint_velocity}
\end{equation}
For the straight-path target $u_t=\Y^{(\mathrm{hr})}-\X_{0,\mathrm{tr}}$, this gives
\begin{equation}
	v_{\phi}(\X_t,t)-u_t
	=\frac{\widehat{\Y}_t-\Y^{(\mathrm{hr})}}{1-t}.
	\label{eq:endpoint_velocity_error}
\end{equation}
Squared velocity matching is therefore equivalent to endpoint regression with weight $(1-t)^{-2}$. We train the endpoint predictor with focal loss, which weights correspondence classification errors by prediction confidence. Equation~\plaineqref{eq:app_endpoint_velocity} defines the sampling velocity. At $t_{K_{\mathrm{II}}-1}=1-\Delta t$, Eq.~\plaineqref{eq:fm_euler} becomes
\begin{equation}
	\X_{K_{\mathrm{II}}}
	=\clip\!\left(\widehat{\Y}_{K_{\mathrm{II}}-1},0,1\right),
\end{equation}
The last step reaches the predicted endpoint without evaluating the denominator at $t=1$.

\paragraph{Conditional Brownian drift identity and initialization.}
For a sampled endpoint pair and its injected noise, substituting $\widehat{\Y}_t=\Y^{(\mathrm{hr})}$ and $\widehat{\bm\epsilon}_t=\bm\epsilon$ into Eqs.~\plaineqref{eq:sb_velocity}--\plaineqref{eq:sb_drift} yields
\begin{align}
	b_{\phi}
	&=\Y^{(\mathrm{hr})}-\X_{0,\mathrm{tr}}
	-\sigma_B\frac{t}{s_t}\bm\epsilon \\
	&=\frac{\Y^{(\mathrm{hr})}-\X_t}{1-t},
	\label{eq:drift_target_direction}
\end{align}
The equality follows from $s_t^2=t(1-t)$ and gives the endpoint-conditioned Brownian drift. At $t=\varepsilon$, the exact conditional mean is $(1-\varepsilon)\X_0+\varepsilon\Y$. Inference omits the unavailable shift $\varepsilon(\Y-\X_0)$ while retaining the noise scale $\sigma_Bs_{\varepsilon}$ in Eq.~\plaineqref{eq:sb_initialization}.

\section{Additional Experiments and Evaluation Protocols}

\subsection{Evaluation Protocol Details}
\label{sec:eval_details}

We follow the Lepard protocol~\cite{li2022lepard} on 4DMatch/4DLoMatch and the filtered 4DMatch-F/4DLoMatch-F splits used by NDP and GraphSCNet~\cite{li2022non,qin2023deep} for downstream registration. At a $4\,\mathrm{cm}$ threshold, NFMR evaluates warped metric points obtained by three-neighbor interpolation from correspondence anchors. IR measures the matches consistent with ground-truth scene flow. For CAPE and DeepDeform, absolute and relative errors are $e_i=\|\widehat{\mathbf f}_i-\mathbf f_i\|_2$ and $r_i=e_i/\max(\|\mathbf f_i\|_2,\epsilon)$. EPE averages $e_i$ in meters. AccS/AccR count points below either the absolute or relative threshold: $(2\,\mathrm{cm},5\%)/(5\,\mathrm{cm},10\%)$ for CAPE and $(2.5\,\mathrm{cm},2.5\%)/(5\,\mathrm{cm},5\%)$ for DeepDeform. OR counts points with $r_i>30\%$. All metrics are averaged over pairs; all except EPE are percentages. Lower EPE/OR and higher NFMR/IR/AccS/AccR are better. Bold and underlined entries mark the best and second-best values within the stated comparison group.

MPC-CAPE and MPC-DD are the public preprocessed CAPE and DeepDeform datasets released with SyNoRiM~\cite{huang2022multiway}. Following RGGT~\cite{zhengrggt}, we expand each of the 209 four-frame MPC-CAPE samples into 12 ordered source--target pairs, giving 2,508 pairs. Diff-Reg and our models use 4DMatch checkpoints and the same pretrained GraphSCNet solver, without CAPE adaptation. The deformation-stratified evaluation reuses these checkpoints without retraining.

For MPC-DD, we fit a least-squares rigid transform to points marked visible by the flow mask, following GraphSCNet~\cite{qin2023deep}. We evaluate the mean residual over the complete source cloud and remove pairs below 1\,cm. Public MPC-DD preprocessing leaves 1,221 pairs, distinct from GraphSCNet's 1,011-pair split. The re-evaluated zero-shot models and their GraphSCNet solver are trained on 4DMatch, with no DeepDeform fine-tuning. EPE is averaged over all source points and then over pairs. SyNoRiM~\cite{huang2022multiway} supplies a target-domain-trained reference.

\subsection{Baseline Training and Comparison Context}
\label{sec:baseline_context}

\paragraph{Result sources.}
The 4DMatch/4DLoMatch correspondence and 4DMatch-F/4DLoMatch-F registration baselines are taken from the cited literature. Baseline rows labeled GraphSCNet report published results using this deformation solver. On CAPE, we evaluate Diff-Reg and our models and quote other scores from RGGT~\cite{zhengrggt}. On DeepDeform, SyNoRiM, Lepard, Diff-Reg, and our models are evaluated on the same 1,221 pairs. The cited AniSym-Net and GeoTransformer+GraphSCNet results use different evaluation protocols.

RGGT~\cite{zhengrggt} provides two 4DMatch-F settings. Compared with RGGT without TRELLIS features, the stochastic bridge lowers EPE by $0.003$, raises AccS/AccR by $12.0/7.4$ points, and lowers the outlier rate by $3.9$ points. With TRELLIS features, RGGT has lower EPE and outlier rate and higher AccR, while our stochastic model has slightly higher AccS ($74.1$ versus $73.9$). RGGT without TRELLIS features does not report 4DLoMatch-F results.

For CAPE, the cited SyNoRiM~\cite{huang2022multiway} result uses target-domain self-supervision and four-view synchronization. GeoTransformer~\cite{qin2022geometric} is trained on ModelNet, and GraphSCNet~\cite{qin2023deep} on 4DMatch. Diff-Reg and our models use 4DMatch checkpoints with pairwise input and no target-domain adaptation.

\paragraph{DeepDeform references.}
Re-evaluated on the same pairs with each directed pair supplied as a two-view sequence, the MPC-DD-trained SyNoRiM reaches $0.0268$ EPE, $72.9$ AccS, $91.8$ AccR, and $12.8$ OR. AniSym-Net~\cite{wang2025anisym} reports $2.66$\,cm EPE and $89.43\%$ AccR in its Table~IV after MPI-FAUST-to-DeepDeform fine-tuning. These correspond to $0.0266$\,m and $89.4\%$ after rounding; AccS and OR are not reported for this setting. AniSym-Net and the published GeoTransformer+GraphSCNet result use different evaluation protocols and are listed separately from the matched zero-shot results.

\subsection{Hierarchical Ablations}
\label{sec:hierarchical_ablation_details}

We train the stochastic coarse-only and high-resolution-only models independently, removing the inactive solver and its loss. Coarse-only lifts the Stage-I prediction for evaluation; high-resolution-only trains Stage~II from an unstructured uniform source.

\begin{table*}[t]
	\caption{Independently trained resolution ablations for the stochastic bridge on 4DLoMatch. The inactive solver and its loss are removed. Results are averaged over three sampling seeds; bold and underlined values denote the best and second-best results.}
	\label{tab:ablation_plan}
	\centering
	\resizebox{0.7\linewidth}{!}{
		\begin{tabular}{lccrr}
			\toprule
			Protocol & Coarse & High-res. & NFMR$\uparrow$ & IR$\uparrow$\\
			\midrule
			Coarse-only (retrained) & $\checkmark$ & --            & 34.04 & 18.03\\
			High-res.-only (retrained) & --           & $\checkmark$  & \underline{80.27} & \underline{75.98}\\
			Full two-stage & $\checkmark$ & $\checkmark$ & \textbf{82.41} & \textbf{78.79}\\
			\bottomrule
	\end{tabular}}
\end{table*}

The full model improves NFMR/IR by $48.37/60.76$ points over coarse-only and $2.14/2.81$ points over high-resolution-only (Table~\ref{tab:ablation_plan}). The following experiments keep trained weights fixed and vary candidate support, source initialization, and inference settings.

\subsection{Inference-Only \texorpdfstring{Top-$K$}{Top-K} Candidate Diagnostic}
\label{sec:topk_candidate_diagnostic}

\paragraph{Protocol.}
This test restricts candidates at inference using one SDE checkpoint trained with all candidates. On 4DLoMatch, we compare full-matrix inference (FULL), an all-candidate mask (ALL), and global coarse-pair Top-$K$ masks for $K\in\{32,64,128,256\}$. We rank all valid coarse pairs by Stage-I score and break ties by flattened index. The selected pairs are lifted to a fixed fine-level mask. Restricted runs apply this mask to the source, stochastic updates, and matching logits, while keeping the dustbin. 

All settings use five coarse steps, twenty fine steps (21 Stage-II NFE), threshold $0.2$, and seeds $0,1,2$. We pair inputs, coarse predictions, and stochastic streams across settings and keep the learned weights fixed.

\paragraph{GT coverage and FULL recovery.}
Let $G$ denote the set of supervised fine-level ground-truth pairs, $M_K$ the lifted candidate mask, and $P_{\mathrm{FULL}}$ the discrete correspondences predicted by unrestricted inference. We measure
\[
\begin{aligned}
\operatorname{Coverage}(K)&=\frac{|G\cap M_K|}{|G|},\\
\operatorname{Recovery}(K)&=\frac{|P_{\mathrm{FULL}}\cap(G\setminus M_K)|}{|G\setminus M_K|}.
\end{aligned}
\]
Recovery counts exact supervised pairs predicted by FULL outside the mask; NFMR/IR instead use geometric tolerance. For each seed, coverage and recovery are calculated from counts summed over test pairs, then averaged across seeds. NFMR/IR are averaged over pairs and seeds. Recovery is undefined when the mask excludes no ground-truth pairs.

\begin{table*}[t]
\caption{Candidate exclusion and recovery on 4DLoMatch for one fixed checkpoint trained with full candidates. Masks are applied at inference. All values are percentages except the FULL-minus-restricted NFMR difference in percentage points (pp). Results average three sampling seeds. FULL recovery is the fraction of excluded GT pairs predicted correctly by unrestricted inference.}
\label{tab:topk_candidate_diagnostic}
\centering
\small
\setlength{\tabcolsep}{5pt}
\begin{tabular}{lrrrrr}
\toprule
Mode & NFMR $\uparrow$ & IR $\uparrow$ & GT coverage & FULL recovery & $\Delta$NFMR (pp) \\
\midrule
FULL & 82.04 & 78.46 & 100.00 & --- & 0.00 \\
ALL & 82.04 & 78.46 & 100.00 & --- & 0.00 \\
Top-32 & 18.19 & 42.88 & 7.46 & 52.66 & 63.85 \\
Top-64 & 28.15 & 40.78 & 13.03 & 51.91 & 53.89 \\
Top-128 & 39.26 & 39.04 & 21.83 & 50.78 & 42.78 \\
Top-256 & 48.94 & 39.26 & 34.23 & 49.30 & 33.10 \\
\bottomrule
\end{tabular}
\end{table*}

\paragraph{Results.}
FULL and ALL give identical scores (Table~\ref{tab:topk_candidate_diagnostic}). Top-256 retains $34.23\%$ of supervised pairs. FULL recovers $49.30\%$ of the pairs excluded by that mask and raises NFMR by $33.10$ points. With smaller masks, FULL recovers $50.78$--$52.66\%$ of the excluded pairs. Full-matrix refinement therefore recovers many correct pairs that coarse selection would discard.

All settings use dense computation and the same full-candidate-trained weights. The masks determine which candidates participate in inference.

\subsection{Inference-Only Source Replacement}
\label{sec:inference_source_probe}

Table~\ref{tab:inference_source_probe} compares three inference settings on 4DLoMatch. The first two use the standard inference procedures of the high-resolution-only and full stochastic models. The third keeps the full-model checkpoint and initializes Stage~II with an independent $\mathcal{U}(0,1)$ matrix, bypassing Stage~I and lifting. All three settings use 20 SDE steps and three seeds.

\begin{table*}[t]
	\caption{Inference-only source replacement for the stochastic bridge on 4DLoMatch. ``Lifted'' denotes the structured Stage-I solution, while $\mathcal{U}(0,1)$ denotes an unstructured high-resolution source. Bold and underlined values denote the best and second-best results.}
	\label{tab:inference_source_probe}
	\centering
	\resizebox{0.6\linewidth}{!}{
		\begin{tabular}{llrr}
			\toprule
			Training protocol & Test-time source & NFMR$\uparrow$ & IR$\uparrow$\\
			\midrule
			High-res.-only & $\mathcal{U}(0,1)$ & 80.27 & 75.98\\
			Full two-stage & Lifted Stage-I & \underline{82.41} & \underline{78.79}\\
			Full two-stage & $\mathcal{U}(0,1)$ & \textbf{82.43} & \textbf{78.92}\\
			\bottomrule
	\end{tabular}}
\end{table*}

At 20 steps, replacing the lifted source with a uniform matrix changes full-model NFMR/IR by $0.02/0.13$ points (Table~\ref{tab:inference_source_probe}). The uniform-source sampling standard deviations are $0.24/0.22$. With the same uniform initialization, the full model exceeds the separately trained high-resolution-only model by $2.16/2.94$ points. The model trained with both stages retains its advantage when the inference source is replaced.

\paragraph{Source sensitivity across inference budgets.}
We next compare lifted and independent $\mathcal{U}(0,1)$ sources at $K\in\{1,2,5,20\}$ Stage-II steps using one fixed SDE checkpoint on 4DLoMatch. The lifted setting uses five Stage-I DDIM steps; uniform initialization bypasses Stage~I. Within each pair and seed, each source is held fixed across budgets and the two settings use matched Stage-II random-number streams at each budget. We use seeds $0,1,2$ and matching threshold $0.2$. Including the final endpoint prediction, the number of Stage-II function evaluations (NFE) is $K+1$.

\begin{table*}[t]
\caption{Source and noise-correction diagnostics across SDE budgets on 4DLoMatch. Values are mean $\pm$ population standard deviation over three sampling seeds. The noise-correction ablation sets the predicted noise to zero during inference while retaining stochastic increments and the final endpoint prediction. Bold marks the higher mean between sources within each correction setting and budget.}
\label{tab:source_budget_probe}
\centering
\small
\setlength{\tabcolsep}{4pt}
\begin{tabular}{rrlrrrr}
\toprule
& & & \multicolumn{2}{c}{With noise correction} & \multicolumn{2}{c}{Without noise correction} \\
\cmidrule(lr){4-5}\cmidrule(lr){6-7}
Steps & NFE & Source & NFMR (\%) $\uparrow$ & IR (\%) $\uparrow$ & NFMR (\%) $\uparrow$ & IR (\%) $\uparrow$ \\
\midrule
1 & 2 & Lifted & $\mathbf{82.79}\pm0.01$ & $\mathbf{79.34}\pm0.03$ & $\mathbf{82.50}\pm0.02$ & $\mathbf{79.17}\pm0.04$ \\
1 & 2 & Uniform & $82.10\pm0.15$ & $79.00\pm0.11$ & $81.73\pm0.29$ & $78.70\pm0.23$ \\
\midrule
2 & 3 & Lifted & $83.06\pm0.15$ & $79.45\pm0.18$ & $\mathbf{81.98}\pm0.17$ & $\mathbf{78.56}\pm0.19$ \\
2 & 3 & Uniform & $\mathbf{83.07}\pm0.22$ & $\mathbf{79.77}\pm0.18$ & $81.02\pm0.13$ & $77.79\pm0.06$ \\
\midrule
5 & 6 & Lifted & $82.15\pm0.07$ & $78.53\pm0.07$ & $\mathbf{81.74}\pm0.06$ & $\mathbf{78.29}\pm0.06$ \\
5 & 6 & Uniform & $\mathbf{82.54}\pm0.13$ & $\mathbf{79.10}\pm0.10$ & $81.19\pm0.15$ & $77.90\pm0.05$ \\
\midrule
20 & 21 & Lifted & $82.00\pm0.06$ & $78.40\pm0.04$ & $\mathbf{81.65}\pm0.17$ & $\mathbf{78.16}\pm0.11$ \\
20 & 21 & Uniform & $\mathbf{82.56}\pm0.15$ & $\mathbf{79.07}\pm0.17$ & $81.36\pm0.21$ & $78.03\pm0.18$ \\
\bottomrule
\end{tabular}
\end{table*}

\paragraph{Results and interpretation.}
With noise correction enabled, the lifted-minus-uniform NFMR/IR differences are $+0.70/+0.34$, $-0.01/-0.32$, $-0.39/-0.57$, and $-0.56/-0.67$ percentage points at $1$, $2$, $5$, and $20$ updates (Table~\ref{tab:source_budget_probe}). Differences are paired by input and seed before rounding. Lifting helps at one update; at larger budgets, uniform initialization matches or exceeds its accuracy. Both sources peak at two updates.

The source sweep uses a fixed checkpoint; the resolution ablations in Table~\ref{tab:ablation_plan} use separately trained models. Runtime for lifted initialization includes the Stage-I computation.

\paragraph{Effect of noise correction.}
The learned noise correction improves SDE refinement for both sources at every tested budget. At two lifted-source updates, it raises NFMR/IR by $1.08/0.89$ points. The ablation sets the predicted noise to zero, reducing the drift to $\hat{\Y}_t-\X_0$, while preserving the initial perturbation, Brownian increments, and final endpoint prediction.

With the correction disabled, lifted-source NFMR falls from $82.50$ at one step to $81.65$ at twenty steps. Without correction, lifted initialization outperforms uniform initialization at every tested budget. Accuracy decreases with more steps both with and without noise correction. The update rule also changes the relative performance of the two sources.

\paragraph{Source replacement in endpoint-CFM.}
The endpoint-CFM sweep uses one checkpoint trained with the lifted-source protocol. Both settings run the five-step Stage-I solver with matched coarse random-number streams; in the uniform setting, only the Stage-II source is replaced by $\mathcal{U}(0,1)$. We verify equal coarse predictions across paired cases and keep each source fixed across budgets. Seeds are $0,1,2$, the matching threshold is $0.2$, and the weights are unchanged. Euler integration is deterministic, so seed variation comes from initialization. The ODE solver uses $K$ Stage-II function evaluations for $K$ integration steps.

\begin{table*}[t]
\caption{Inference-time source replacement for one endpoint-CFM model trained with the lifted-source protocol on 4DLoMatch. Both settings share the same checkpoint. Values are mean $\pm$ population standard deviation across three initialization seeds. Bold indicates the higher mean within each budget. Stage-II NFE equals the number of steps.}
\label{tab:source_budget_ode}
\centering
\begin{tabular}{rlrr}
\toprule
Steps / NFE & Source & NFMR (\%) $\uparrow$ & IR (\%) $\uparrow$ \\
\midrule
1 & Lifted & $\mathbf{75.82}\pm0.10$ & $\mathbf{70.07}\pm0.16$ \\
1 & Uniform & $70.95\pm0.07$ & $65.68\pm0.06$ \\
\midrule
2 & Lifted & $76.32\pm0.01$ & $70.04\pm0.07$ \\
2 & Uniform & $\mathbf{76.76}\pm0.04$ & $\mathbf{70.71}\pm0.05$ \\
\midrule
5 & Lifted & $77.80\pm0.06$ & $71.05\pm0.04$ \\
5 & Uniform & $\mathbf{78.04}\pm0.12$ & $\mathbf{71.59}\pm0.12$ \\
\midrule
20 & Lifted & $78.44\pm0.10$ & $71.57\pm0.10$ \\
20 & Uniform & $\mathbf{78.57}\pm0.11$ & $\mathbf{71.86}\pm0.11$ \\
\bottomrule
\end{tabular}
\end{table*}

\paragraph{Inference-time initialization sensitivity.}
For endpoint-CFM, the lifted-minus-uniform NFMR/IR differences are $+4.87/+4.40$, $-0.44/-0.68$, $-0.23/-0.54$, and $-0.12/-0.29$ percentage points at $1$, $2$, $5$, and $20$ updates (Table~\ref{tab:source_budget_ode}), computed before rounding. The lifted source helps most at one update. Uniform initialization gives slightly higher means with two or more updates, and both sources improve from two to five to twenty updates.

\paragraph{Noise correction and solver comparison.}
The SDE design achieves $82.00/78.40$ NFMR/IR at twenty lifted-source updates, compared with $78.44/71.57$ for endpoint-CFM. Its learned noise correction improves the SDE result from $81.65/78.16$ to $82.00/78.40$. The independently trained ODE and SDE models implement their respective paths, prediction heads, losses, and inference rules; the correction ablation evaluates the SDE drift within the same model.

\subsection{Solver-Step Sensitivity}
\label{sec:solver_step_appendix}

Table~\ref{tab:solver_steps} studies the iteration budgets of both Stage-II models. Let $K_{\mathrm{I}}$ denote the number of Stage-I DDIM steps at the coarse resolution, and let $K_{\mathrm{II}}$ denote the number of Stage-II Euler or Euler--Maruyama steps at the high resolution. These inference-only tests reuse fixed checkpoints without retraining. We first vary $K_{\mathrm{I}}\in\{1,3,5,10,20\}$ while fixing $K_{\mathrm{II}}=20$. We then vary $K_{\mathrm{II}}\in\{1,2,5,10,20,40\}$ while fixing $K_{\mathrm{I}}=5$. We evaluate endpoint-CFM once for each pair and average the Brownian-bridge results over three seeds.

\paragraph{Runtime and memory protocol.}
We measure synchronized wall-clock latency on an NVIDIA RTX 4090 GPU with a batch size of one and average it over all test pairs. The backbone features are computed once for each pair and then reused, while the matching transformer and prediction heads run at every step. $T_{\mathrm{I}}$ includes backbone encoding, multiresolution preprocessing, coarse DDIM inference, and hierarchical lifting. $T_{\mathrm{II}}$ includes high-resolution bridge integration, correspondence extraction, and Soft Procrustes. Both measurements exclude data loading, metric computation, and downstream RANSAC. Peak GPU memory includes model parameters and intermediate tensors but excludes the reserved CUDA cache. We average the stochastic latency across seeds and report the largest memory peak.

\begin{table*}[t]
	\caption{Step budgets and computational cost under standard ODE and SDE sampling on 4DLoMatch. SDE sampling retains the Brownian increment at each update. The Stage-I sweep fixes $K_{\mathrm{II}}=20$, and the Stage-II sweep fixes $K_{\mathrm{I}}=5$. Brownian-bridge results are mean$\pm$standard deviation over three sampling seeds. $T_{\mathrm{I}}$ and $T_{\mathrm{II}}$ are measured separately per pair on the same hardware. For correspondence metrics, bold and underlined values denote the best and second-best settings within each sweep and branch; latency and memory are not ranked.}
	\label{tab:solver_steps}
	\centering
	\resizebox{\textwidth}{!}{
		\begin{tabular}{lccrrrrrrrrr}
			\toprule
			&&&
			\multicolumn{2}{c}{Ours$^{\mathrm{ODE}}$} &
			\multicolumn{2}{c}{Ours$^{\mathrm{Stochastic}}$} &
			\multicolumn{3}{c}{Latency (ms)$\downarrow$} &
			\multicolumn{2}{c}{Memory (GiB)$\downarrow$} \\
			\cmidrule(lr){4-5}
			\cmidrule(lr){6-7}
			\cmidrule(lr){8-10}
			\cmidrule(lr){11-12}
			Sweep & $K_{\mathrm{I}}$ & $K_{\mathrm{II}}$ &
			NFMR$\uparrow$ & IR$\uparrow$ &
			NFMR$\uparrow$ & IR$\uparrow$ &
			$T_{\mathrm{I}}^{\mathrm{ODE}}/T_{\mathrm{I}}^{\mathrm{SDE}}$ & $T_{\mathrm{II}}^{\mathrm{ODE}}$ & $T_{\mathrm{II}}^{\mathrm{SDE}}$ &
			ODE & SDE \\
			\midrule
			Stage I & 1  & 20 & 78.27 & 71.42 & $\underline{82.12\pm0.08}$ & $78.47\pm0.08$ & 50/58 & 578 & 727 & 1.86 & 1.86 \\
			& 3  & 20 & 78.36 & 71.52 & $82.03\pm0.03$ & $78.40\pm0.06$ & 114/96 & 748 & 664 & 1.85 & 1.86 \\
			& 5  & 20 & \underline{78.46} & \underline{71.67} & $82.04\pm0.08$ & $78.46\pm0.09$ & 162/133 & 740 & 639 & 1.85 & 1.86 \\
			& 10 & 20 & 78.37 & 71.51 & $\mathbf{82.19\pm0.20}$ & $\mathbf{78.58\pm0.21}$ & 283/194 & 727 & 503 & 1.86 & 1.86 \\
			& 20 & 20 & \textbf{78.68} & \textbf{71.80} & $\underline{82.12\pm0.04}$ & $\underline{78.54\pm0.02}$ & 513/308 & 713 & 426 & 1.85 & 1.86 \\
			\midrule
			Stage II & 5 & 1  & 75.76 & 70.11 & $\underline{82.94\pm0.10}$ & $\underline{79.52\pm0.04}$ & 162/99 & 46 & 44 & 1.85 & 1.85 \\
			& 5 & 2  & 76.38 & 70.11 & $\mathbf{83.16\pm0.13}$ & $\mathbf{79.62\pm0.11}$ & 162/98 & 82 & 64 & 1.86 & 1.85 \\
			& 5 & 5  & \underline{77.98} & 71.23 & $82.29\pm0.03$ & $78.73\pm0.03$ & 159/96 & 189 & 125 & 1.85 & 1.85 \\
			& 5 & 10 & \textbf{78.46} & \textbf{71.71} & $82.22\pm0.11$ & $78.61\pm0.08$ & 152/94 & 355 & 225 & 1.85 & 1.85 \\
			& 5 & 20 & \textbf{78.46} & \underline{71.67} & $82.04\pm0.08$ & $78.46\pm0.09$ & 162/133 & 740 & 639 & 1.85 & 1.86 \\
			& 5 & 40 & \textbf{78.46} & 71.63 & $82.03\pm0.21$ & $78.44\pm0.18$ & 150/89 & 1380 & 814 & 1.85 & 1.86 \\
			\bottomrule
	\end{tabular}}
\end{table*}

Increasing Stage-I steps from 1 to 20 changes ODE NFMR/IR by $0.41/0.38$ points and SDE by $0.00/0.07$ points (Table~\ref{tab:solver_steps}). Varying the fine-stage budget has a larger effect.

ODE iterative refinement improves matching beyond the single-step endpoint prediction. One Euler step at $t=0$ reaches the clipped endpoint prediction; ten steps raise NFMR/IR from $75.76/70.11$ to $78.46/71.71$ ($+2.70/+1.60$ points), after which accuracy plateaus. The stochastic model improves from $82.94/79.52$ at one update to $83.16/79.62$ at two ($+0.22/+0.10$ points), then loses accuracy at larger budgets. Each stochastic run includes a final endpoint prediction: $K$ updates cost $K+1$ network evaluations, compared with $K$ for endpoint-CFM. Both models use twenty updates by default.

\paragraph{Runtime and memory.}
Ten ODE steps take $355$\,ms in Stage~II, compared with $740$\,ms for twenty, a $2.08\times$ speedup. NFMR/IR are $78.46/71.71$ and $78.46/71.67$, respectively. Two SDE updates take $64$\,ms, compared with $639$\,ms for twenty, a $9.98\times$ speedup, and improve NFMR/IR from $82.04/78.46$ to $83.16/79.62$. Peak memory stays within $1.85$--$1.86$\,GiB. Stage-I costs are reported separately in Table~\ref{tab:solver_steps}.

\subsection{Inference Controls on a Fixed SDE Checkpoint}
\label{sec:sde_inference_controls}

\paragraph{Protocol.}
We evaluate one frozen SDE-trained checkpoint on 4DLoMatch. The lifted source uses five Stage-I DDIM updates; the uniform source draws an independent $\mathcal U(0,1)$ matrix and bypasses Stage~I. Within each source, all modes and budgets use the same input and initial state $\X_{\varepsilon}=\X_0+\sigma_B\sqrt{\varepsilon(1-\varepsilon)}\,\mathbf z$, with $\sigma_B=0.1$ and $\varepsilon=0.02$. Correspondences use the same mutual matching rule and confidence threshold $0.2$. The subsequent updates are deterministic conditional on this initial state.

\paragraph{Update rules.}
The forward-drift mode retains the endpoint and noise heads and integrates $\X_{k+1}=\X_k+\Delta t\,b_\phi(\X_k,t_k)$, using the complete learned drift in Eq.~\plaineqref{eq:sb_drift}. It removes only the Brownian increment from Eq.~\plaineqref{eq:euler_maruyama}. 

The probability-flow mode integrates $\X_{k+1}=\X_k+\Delta t\,v_\phi(\X_k,t_k)$ using Eq.~\plaineqref{eq:sb_velocity}. Both run $K$ updates over $[\varepsilon,1-\varepsilon]$ and read the endpoint once more, costing $K+1$ Stage-II network evaluations. 

Direct endpoint prediction queries the endpoint head once at $(\X_{\varepsilon},\varepsilon)$ and uses its output for matching. All three modes use the same weights; the probability-flow mode here differs from the separately trained endpoint-CFM model. Code names are \texttt{drift\_no\_noise}, \texttt{pf\_ode}, and \texttt{endpoint}, respectively.

\begin{table*}[t]
 \caption{Inference-rule ablation on a fixed SDE checkpoint, 4DLoMatch. NFMR and IR are percentages. Initial perturbations are paired within each source; neither update mode injects subsequent random increments. Direct endpoint prediction is evaluated once per source, at 1 NFE.}
 \label{tab:sde_inference_full}
 \centering\small
 \setlength{\tabcolsep}{6pt}
 \begin{tabular}{llrrrr}
  \toprule
  Source & Inference mode & Updates & Stage-II NFE & NFMR$\uparrow$ & IR$\uparrow$\\
  \midrule
  Lifted & Forward drift & 1 & 2 & 83.94 & 80.49\\
  Lifted & Forward drift & 2 & 3 & 83.44 & 79.84\\
  Lifted & Forward drift & 5 & 6 & 82.62 & 79.03\\
  Lifted & Forward drift & 20 & 21 & 82.37 & 78.70\\
  Lifted & Probability flow & 1 & 2 & 74.06 & 72.56\\
  Lifted & Probability flow & 2 & 3 & 75.32 & 73.86\\
  Lifted & Probability flow & 5 & 6 & 78.05 & 75.97\\
  Lifted & Probability flow & 20 & 21 & 80.04 & 76.89\\
  Lifted & Direct endpoint & 0 & 1 & 79.28 & 76.69\\
  \midrule
  Uniform & Forward drift & 1 & 2 & 82.82 & 79.61\\
  Uniform & Forward drift & 2 & 3 & 82.56 & 79.33\\
  Uniform & Forward drift & 5 & 6 & 83.16 & 79.64\\
  Uniform & Forward drift & 20 & 21 & 82.63 & 79.12\\
  Uniform & Probability flow & 1 & 2 & 74.85 & 73.02\\
  Uniform & Probability flow & 2 & 3 & 74.76 & 72.96\\
  Uniform & Probability flow & 5 & 6 & 76.42 & 74.42\\
  Uniform & Probability flow & 20 & 21 & 80.01 & 77.41\\
  Uniform & Direct endpoint & 0 & 1 & 75.12 & 72.79\\
  \bottomrule
 \end{tabular}
\end{table*}

\paragraph{Target-directed contraction and learned-drift error.}
The endpoint-conditioned drift in Eq.~\plaineqref{eq:drift_target_direction} defines the reference field $b^*(\X,t)=(\Y-\X)/(1-t)$. An Euler update with this field and step size $0<h\leq 1-t$ contracts the target error:
\begin{equation}
 \X_{t+h}^{*,\mathrm{Euler}}-\Y
 =\left(1-\frac{h}{1-t}\right)(\X_t-\Y).
 \label{eq:drift_error_contraction}
\end{equation}
For the implemented initialization $\X_\varepsilon=\X_0+\sigma_Bs_\varepsilon\mathbf z$, define $\Delta b_\varepsilon=b_\phi(\X_\varepsilon,\varepsilon)-(\Y-\X_\varepsilon)/(1-\varepsilon)$. The single update $\X^+=\X_\varepsilon+(1-2\varepsilon)b_\phi(\X_\varepsilon,\varepsilon)$ then satisfies
\begin{equation}
 \X^+-\Y
 =\frac{\varepsilon}{1-\varepsilon}(\X_\varepsilon-\Y)
 +(1-2\varepsilon)\Delta b_\varepsilon.
 \label{eq:drift_learned_update_error}
\end{equation}
The first term is the contracted initial error; the second is the error contributed by the learned drift. For $\Delta b_\varepsilon=0$, the ideal contraction factor is $\varepsilon/(1-\varepsilon)\approx0.0204$ at $\varepsilon=0.02$. The equation describes the state before the final endpoint prediction at $1-\varepsilon$, which completes the two-NFE procedure evaluated below.

\paragraph{Few-step SDE refinement.}
With the lifted source, one forward-drift update followed by endpoint prediction reaches $83.94/80.49$ NFMR/IR, compared with $79.28/76.69$ for direct endpoint prediction. The gains are $4.66/3.80$ percentage points. With the uniform source, the corresponding gain is $7.70/6.82$ points, from $75.12/72.79$ to $82.82/79.61$. On 4DMatch, the same lifted-source configuration reaches $93.10/92.25$ NFMR/IR at 2 Stage-II NFE, improving over the standard stochastic result by $0.67/0.78$ points.

\paragraph{Forward drift and probability flow.}
Forward-drift updates outperform probability-flow updates at every matched source and NFE budget in Table~\ref{tab:sde_inference_full}. At 2 NFE, the lifted-source gap is $9.88/7.93$ NFMR/IR points. The gap narrows to $2.33/1.81$ points at 21 NFE as probability-flow accuracy increases. Uniform-source probability flow also improves overall with more updates, reaching $80.01/77.41$ at 21 NFE. The largest gap occurs at one update, or two Stage-II network evaluations.

\paragraph{Initialization and budget.}
The lifted source improves direct endpoint NFMR/IR over uniform initialization by $4.16/3.90$ points and retains a $1.12/0.88$-point advantage after one forward-drift update. Lifted-source refinement reaches its highest accuracy at this single update; the uniform source peaks at five updates with $83.16/79.64$ and exceeds lifting at five and twenty updates.

\subsection{CAPE Results by Deformation Magnitude}
\label{sec:cape_deformation_details}

We sort the 2,508 CAPE pairs by mean ground-truth flow magnitude and divide them into three equal groups. The same tertile boundaries are used for every method.

\begin{table*}[t]
	\caption{Zero-shot CAPE performance grouped by mean ground-truth deformation magnitude. EPE is in meters; other metrics are percentages. The groups use shared test-set tertile boundaries and contain 836 pairs each; stochastic results are averaged over three sampling seeds. Best and second-best values are determined separately within each deformation subset.}
	\label{tab:cape_deformation}
	\centering
	\resizebox{0.7\textwidth}{!}{
		\begin{tabular}{llrrrr}
			\toprule
			Method & Deformation & EPE (m)$\downarrow$ & AccS$\uparrow$ & AccR$\uparrow$ & OR$\downarrow$\\
			\midrule
			\multirow{3}{*}{Diff-Reg} & Small  & 0.012240 & 87.304 & \textbf{97.682} & \underline{64.826}\\
			& Medium & 0.017558 & 78.749 & 95.285 & 42.121\\
			& Large  & 0.024673 & 69.102 & 92.249 & 29.312\\
			\midrule
			\multirow{3}{*}{Ours$^{\mathrm{ODE}}$} & Small  & \underline{0.011739} & \underline{87.903} & \underline{97.534} & 65.754\\
			& Medium & \underline{0.016726} & \underline{79.191} & \underline{95.419} & \underline{41.136}\\
			& Large  & \underline{0.022123} & \underline{70.466} & \underline{93.025} & \underline{27.332}\\
			\midrule
			\multirow{3}{*}{Ours$^{\mathrm{Stochastic}}$} & Small  & \textbf{0.011418} & \textbf{88.459} & \textbf{97.682} & \textbf{64.170}\\
			& Medium & \textbf{0.016416} & \textbf{80.369} & \textbf{95.827} & \textbf{39.982}\\
			& Large  & \textbf{0.021271} & \textbf{72.417} & \textbf{93.681} & \textbf{26.081}\\
			\bottomrule
	\end{tabular}}
\end{table*}

\paragraph{Analysis of deformation severity.}
SDE has the lowest EPE and OR and the highest AccS in all three deformation groups (Table~\ref{tab:cape_deformation}). AccR is highest for medium and large deformations and tied for highest for small deformations at the reported precision. On the large-deformation subset, SDE reaches $0.021271$ EPE and $72.417\%$ AccS; ODE reaches $0.022123$ and $70.466\%$. We compare OR within each group because its relative-error denominator depends on motion magnitude.

\subsection{Zero-Shot Transfer to Complete Shapes on TOSCA}
\label{sec:tosca_zero_shot}

We test whether both variants trained on partial scans in 4DMatch transfer to complete-shape correspondence on TOSCA, without training on SMAL or TOSCA. We use 41 animal shapes and all 286 within-category directed, non-self pairs under the DPC/SE-ORNet reference protocol. Each shape is represented by 1,024 points sampled from its complete surface. Corresponding vertex subsets are fixed during benchmark preparation, and coordinates are divided by 100. Ground-truth maps are used only for data preparation and evaluation.

For dense prediction, each input source point inherits the row-maximum match of its nearest high-resolution source anchor, and the matched target anchor is mapped to the nearest input target point. No GraphSCNet or additional test-time optimization is used. Acc@1\% counts predictions whose Euclidean error is below $1\%$ of the sampled target's maximum pairwise distance, following \citet{Chen_2026_CVPR}. 

Both variants use their epoch-30 checkpoints and the same test inputs and decoder. ODE uses 10 coarse and 20 fine steps; the stochastic model uses 5 coarse and 20 fine steps.

\begin{table*}[t]
 \caption{TOSCA correspondence accuracy, grouped by additional test-time optimization. SMAL-trained values are from Table~1 of \citet{Chen_2026_CVPR}; all rows exclude TOSCA training. Ours reports pair-averaged mean $\pm$ population standard deviation over three inference seeds. Training data and test-time optimization are specified for each method.}
 \label{tab:tosca_zero_shot}
 \centering\small
 \setlength{\tabcolsep}{7pt}
 \begin{tabular}{lllcr}
  \toprule
  Method & Venue & Training data & \shortstack{Additional\\optimization} & Acc@1\% ($\%$)$\uparrow$\\
  \midrule
  \multicolumn{5}{l}{\textit{With additional test-time optimization}}\\
  EquiShape+ & TIP 2025 & SMAL & Yes & 57.7\\
  Chen et al.+ & CVPR 2026 & SMAL & Yes & 60.5\\
  \midrule
  \multicolumn{5}{l}{\textit{Without additional test-time optimization}}\\
  CorrNet3D & CVPR 2021 & SMAL & No & 5.3\\
  DPC & 3DV 2021 & SMAL & No & 33.2\\
  SE-ORNet & CVPR 2023 & SMAL & No & 36.4\\
  HSTR & TIP 2023 & SMAL & No & 33.9\\
  TANet & CVPR 2024 & SMAL & No & 37.1\\
  AMIGO & AAAI 2025 & SMAL & No & 38.5\\
  DiffCorr & AAAI 2025 & SMAL & No & $41.6\pm0.6$\\
  DV-Matcher & CVPR 2025 & SMAL & No & 39.5\\
  Chen et al. & CVPR 2026 & SMAL & No & 54.0\\
  \midrule
  \textbf{Ours}$^{\mathrm{ODE}}$ & -- & 4DMatch & No & $49.51\pm0.02$\\
  \textbf{Ours}$^{\mathrm{Stochastic}}$ & -- & 4DMatch & No & $54.01\pm0.02$\\
  \bottomrule
 \end{tabular}
\end{table*}

On complete TOSCA shapes, SDE reaches $54.01\%$ accuracy and ODE reaches $49.51\%$, a gap of $4.50$ points (Table~\ref{tab:tosca_zero_shot}). Both models were trained on partial 4DMatch scans and use no target-domain training or additional test-time optimization.

Our models use supervised 4DMatch training; the reference methods use unsupervised SMAL training and their published evaluation protocols.

\section{Matching-Matrix Trajectory Visualization}
\label{sec:trajectory_visualization}

\begin{figure*}[!t]
	\centering
	\includegraphics[width=0.98\textwidth]{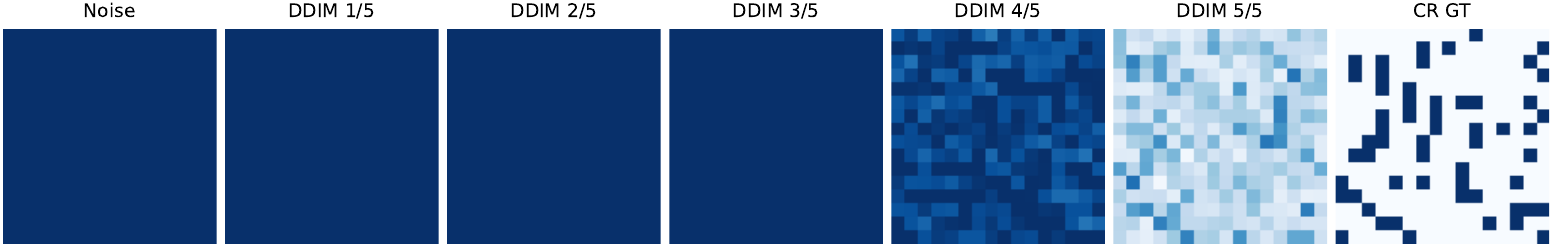}
	\vspace{-1.5mm}

	{\footnotesize (a) Compact coarse-resolution trajectory from endpoint-CFM.}
	\vspace{1mm}

	\includegraphics[width=0.98\textwidth]{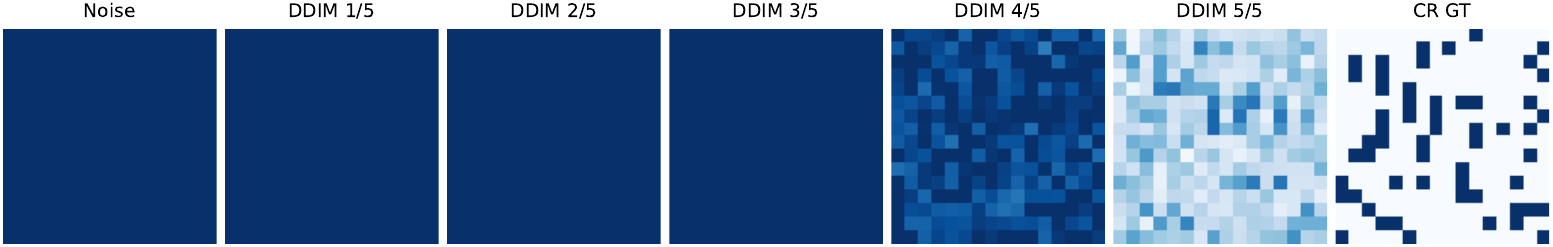}
	\vspace{-1.5mm}

	{\footnotesize (b) Compact coarse-resolution trajectory from the stochastic bridge.}
	\vspace{1mm}

	\includegraphics[width=0.98\textwidth]{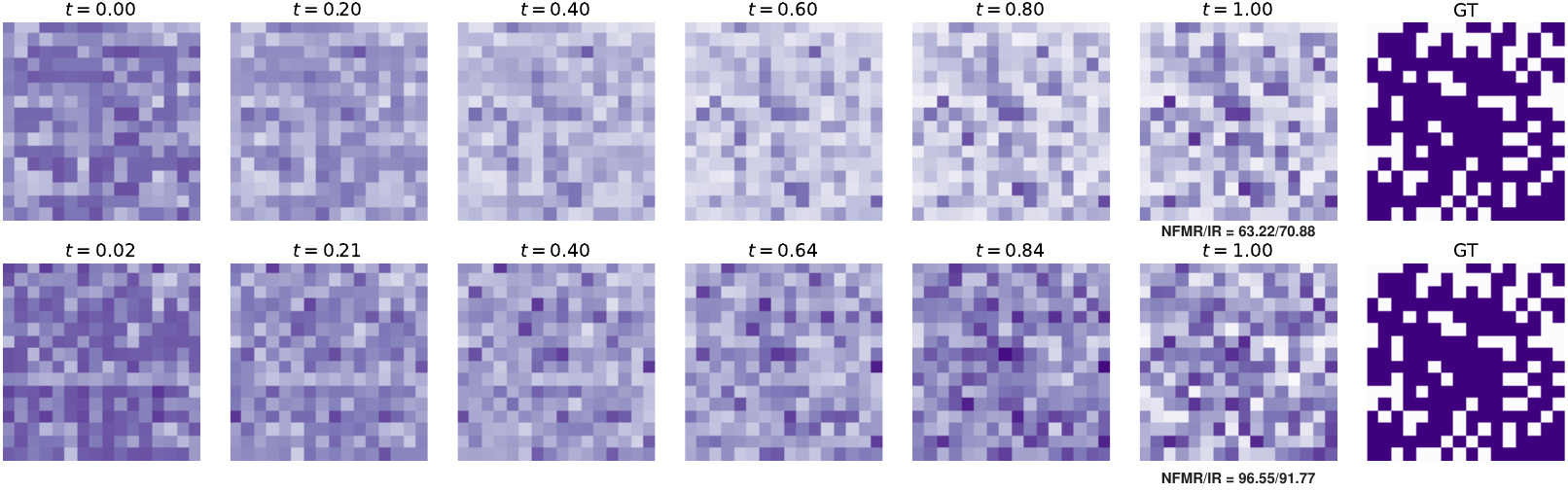}
	\vspace{-1.5mm}

	{\footnotesize (c) Index 00105: high-resolution Stage-II trajectories of endpoint-CFM (top) and the stochastic bridge (bottom).}
	\vspace{1mm}

	\caption{Matching-matrix trajectories on 4DLoMatch. (a)--(b) show the initial state and five Stage-I DDIM updates for index 00105, in the compact coarse-resolution space. (c) compares its deterministic ODE and stochastic SDE trajectories in the high-resolution space; endpoint-CFM obtains 63.22 NFMR and 70.88 IR, and the stochastic bridge obtains 96.55 NFMR and 91.77 IR. Matrices are max-pooled into $16\times16$ mosaics for visualization only; the final column of each panel is the corresponding ground-truth matching matrix.}
	\label{fig:matching_matrix_trajectories}
\end{figure*}

Figure~\ref{fig:matching_matrix_trajectories} visualizes example 00105, on which the stochastic model achieves higher correspondence scores than endpoint-CFM. The coarse panels show the initial state and five DDIM updates; the fine panel shows the Stage-II trajectories. The trajectories differ mainly in Stage~II. Matrices are pooled only for display; inference uses the original resolution.

\section{Failure Cases}
\label{sec:failure_cases}

\begin{figure*}[!t]
	\centering
	\begin{tikzpicture}
		\node[anchor=south west,inner sep=0] (capeFailures) at (0,0)
		{\includegraphics[width=0.98\textwidth]{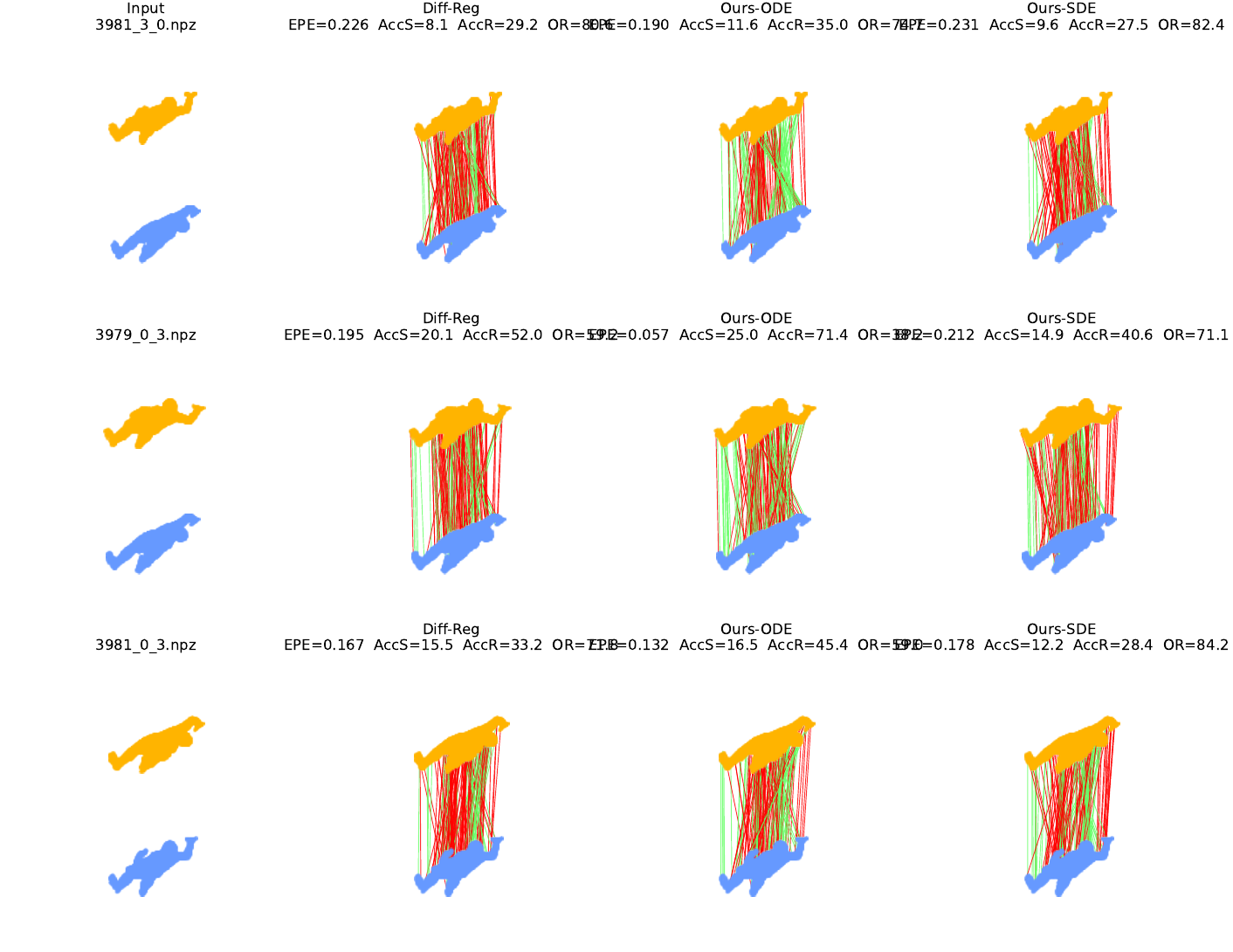}};
		\begin{scope}[x={(capeFailures.south east)},y={(capeFailures.north west)}]
			\foreach \y in {0.97,0.65,0.33}{\fill[white] (0,\y-0.035) rectangle (1,\y+0.035);}
			\node[align=center,font=\fontsize{5.2}{5.6}\selectfont] at (0.125,0.97) {Input\\3981\_3\_0};
			\node[align=center,font=\fontsize{5.2}{5.6}\selectfont] at (0.375,0.97) {Diff-Reg\\EPE=.226, OR=80.6\\AccS=8.1, AccR=29.2};
			\node[align=center,font=\fontsize{5.2}{5.6}\selectfont] at (0.625,0.97) {Ours-ODE\\EPE=.190, OR=74.7\\AccS=11.6, AccR=35.0};
			\node[align=center,font=\fontsize{5.2}{5.6}\selectfont] at (0.875,0.97) {Ours-SDE\\EPE=.231, OR=82.4\\AccS=9.6, AccR=27.5};
			\node[align=center,font=\fontsize{5.2}{5.6}\selectfont] at (0.125,0.65) {Input\\3979\_0\_3};
			\node[align=center,font=\fontsize{5.2}{5.6}\selectfont] at (0.375,0.65) {Diff-Reg\\EPE=.195, OR=59.2\\AccS=20.1, AccR=52.0};
			\node[align=center,font=\fontsize{5.2}{5.6}\selectfont] at (0.625,0.65) {Ours-ODE\\EPE=.057, OR=38.2\\AccS=25.0, AccR=71.4};
			\node[align=center,font=\fontsize{5.2}{5.6}\selectfont] at (0.875,0.65) {Ours-SDE\\EPE=.212, OR=71.1\\AccS=14.9, AccR=40.6};
			\node[align=center,font=\fontsize{5.2}{5.6}\selectfont] at (0.125,0.33) {Input\\3981\_0\_3};
			\node[align=center,font=\fontsize{5.2}{5.6}\selectfont] at (0.375,0.33) {Diff-Reg\\EPE=.167, OR=71.8\\AccS=15.5, AccR=33.2};
			\node[align=center,font=\fontsize{5.2}{5.6}\selectfont] at (0.625,0.33) {Ours-ODE\\EPE=.132, OR=59.0\\AccS=16.5, AccR=45.4};
			\node[align=center,font=\fontsize{5.2}{5.6}\selectfont] at (0.875,0.33) {Ours-SDE\\EPE=.178, OR=84.2\\AccS=12.2, AccR=28.4};
		\end{scope}
	\end{tikzpicture}
	\caption{CAPE failure cases. The three common test pairs with the highest EPE in the displayed Ours-SDE run are shown in descending order. Each row contains the separated input point clouds followed by correspondences from Diff-Reg, Ours-ODE, and Ours-SDE. Orange and blue denote the source and target, while light-green and red lines denote correct and incorrect correspondences, respectively. EPE is measured in meters; AccS, AccR, and OR are percentages. The separation between the source and target is introduced only for visualization.}
	\label{fig:cape_failure_cases}
\end{figure*}

\begin{figure*}[!t]
	\centering
	\begin{tikzpicture}
		\node[anchor=south west,inner sep=0] (deepFailures) at (0,0)
		{\includegraphics[width=0.98\textwidth]{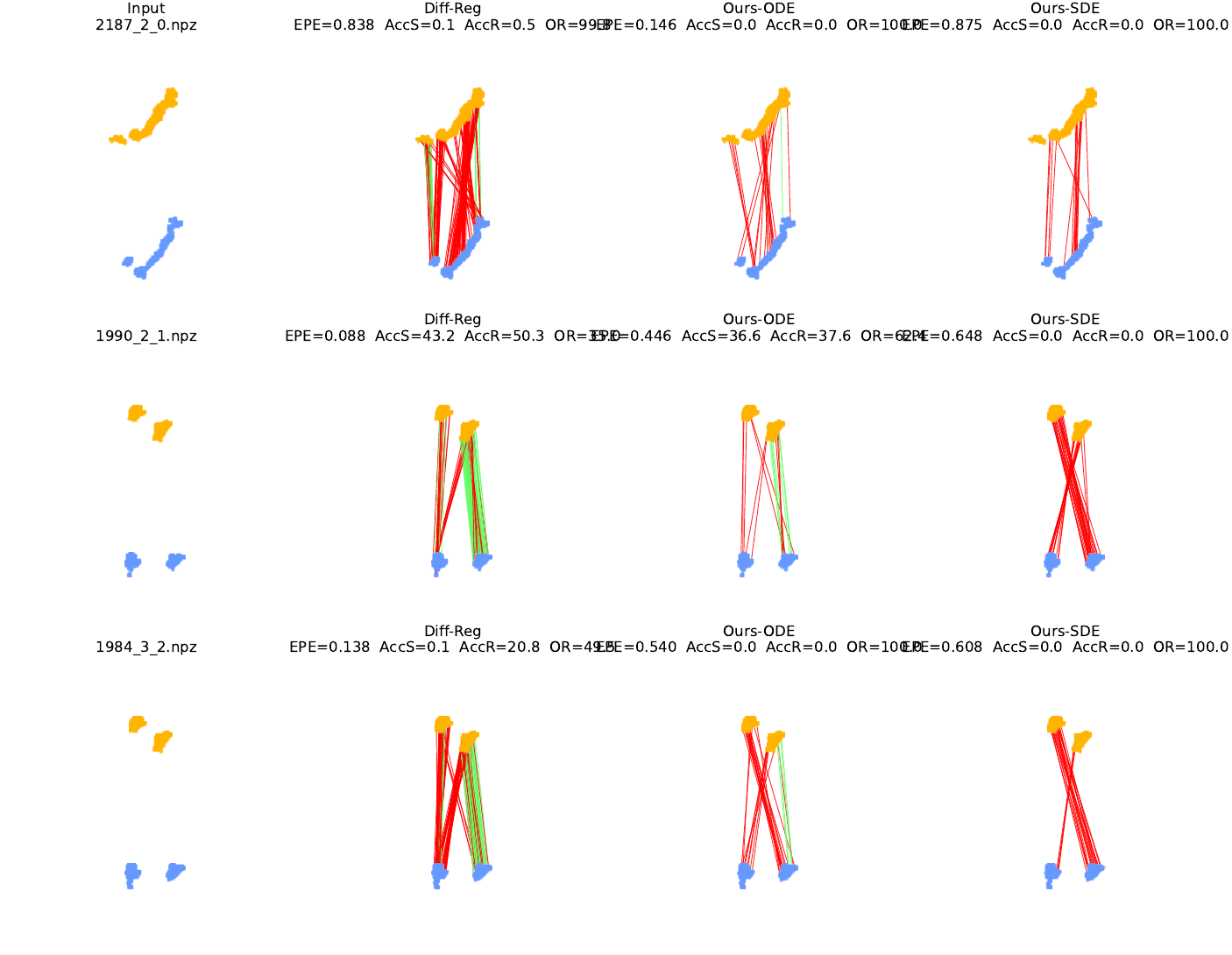}};
		\begin{scope}[x={(deepFailures.south east)},y={(deepFailures.north west)}]
			\foreach \y in {0.97,0.65,0.33}{\fill[white] (0,\y-0.035) rectangle (1,\y+0.035);}
			\node[align=center,font=\fontsize{5.2}{5.6}\selectfont] at (0.125,0.97) {Input\\2187\_2\_0};
			\node[align=center,font=\fontsize{5.2}{5.6}\selectfont] at (0.375,0.97) {Diff-Reg\\EPE=.838, OR=99.8\\AccS=0.1, AccR=0.5};
			\node[align=center,font=\fontsize{5.2}{5.6}\selectfont] at (0.625,0.97) {Ours-ODE\\EPE=.146, OR=100.0\\AccS=0.0, AccR=0.0};
			\node[align=center,font=\fontsize{5.2}{5.6}\selectfont] at (0.875,0.97) {Ours-SDE\\EPE=.875, OR=100.0\\AccS=0.0, AccR=0.0};
			\node[align=center,font=\fontsize{5.2}{5.6}\selectfont] at (0.125,0.65) {Input\\1990\_2\_1};
			\node[align=center,font=\fontsize{5.2}{5.6}\selectfont] at (0.375,0.65) {Diff-Reg\\EPE=.088, OR=35.0\\AccS=43.2, AccR=50.3};
			\node[align=center,font=\fontsize{5.2}{5.6}\selectfont] at (0.625,0.65) {Ours-ODE\\EPE=.446, OR=62.4\\AccS=36.6, AccR=37.6};
			\node[align=center,font=\fontsize{5.2}{5.6}\selectfont] at (0.875,0.65) {Ours-SDE\\EPE=.648, OR=100.0\\AccS=0.0, AccR=0.0};
			\node[align=center,font=\fontsize{5.2}{5.6}\selectfont] at (0.125,0.33) {Input\\1984\_3\_2};
			\node[align=center,font=\fontsize{5.2}{5.6}\selectfont] at (0.375,0.33) {Diff-Reg\\EPE=.138, OR=49.5\\AccS=0.1, AccR=20.8};
			\node[align=center,font=\fontsize{5.2}{5.6}\selectfont] at (0.625,0.33) {Ours-ODE\\EPE=.540, OR=100.0\\AccS=0.0, AccR=0.0};
			\node[align=center,font=\fontsize{5.2}{5.6}\selectfont] at (0.875,0.33) {Ours-SDE\\EPE=.608, OR=100.0\\AccS=0.0, AccR=0.0};
		\end{scope}
	\end{tikzpicture}
	\caption{DeepDeform failure cases under the same protocol as Fig.~\ref{fig:cape_failure_cases}. The rows are the three common test pairs with the highest EPE in the displayed Ours-SDE run. Colors, correspondence correctness, metrics, and the visualization-only separation follow Fig.~\ref{fig:cape_failure_cases}.}
	\label{fig:deepdeform_failure_cases}
\end{figure*}

We examine the three pairs with the highest EPE in the displayed stochastic run on each dataset, restricted to pairs evaluated by all three methods. Figures~\ref{fig:cape_failure_cases} and~\ref{fig:deepdeform_failure_cases} show ambiguous correspondences between similar body regions under partial visibility.

In the CAPE examples (Fig.~\ref{fig:cape_failure_cases}), the lying poses make body regions difficult to distinguish from overall shape. Limb bends and visible surface details provide local cues, but all three methods produce incorrect matches. The stochastic EPE values are $0.231$, $0.212$, and $0.178$, compared with $0.190$, $0.057$, and $0.132$ for endpoint-CFM. Diff-Reg also has lower EPE than the stochastic model on these three pairs.

The DeepDeform examples contain partially visible bodies and similarly shaped local regions (Fig.~\ref{fig:deepdeform_failure_cases}). On the first pair, endpoint-CFM has lower EPE ($0.146$) than Diff-Reg ($0.838$) and the stochastic model ($0.875$), yet its AccS and AccR are zero and OR is $100\%$. The stochastic model has EPEs of $0.875$, $0.648$, and $0.608$ across the three pairs, with zero AccS/AccR and $100\%$ OR throughout. All three methods fail on the first pair.

\section{Limitations and Future Work}

\paragraph{Deterministic transport.}
Endpoint-CFM uses fixed coupling and straight training paths. Reflow or trajectory distillation may reduce the number of inference steps needed to reach a given matching accuracy.

\paragraph{Trajectory learning.}
During training, we sample states analytically using the ground-truth endpoint. During inference, each state depends on earlier predictions. Training on predicted trajectories could help reduce errors that accumulate over many updates.

\end{document}